# Short and Long Supports for Constraint Propagation

**Peter Nightingale**                                    PWN1@ST-ANDREWS.AC.UK
**Ian P. Gent**                                    IAN.GENT@ST-ANDREWS.AC.UK
**Christopher Jefferson**                                    CAJ21@ST-ANDREWS.AC.UK
**Ian Miguel**                                    IJM@ST-ANDREWS.AC.UK
*School of Computer Science, University of St Andrews,*
*St Andrews, Fife KY16 9SX, UK*

## Abstract

Special-purpose constraint propagation algorithms frequently make implicit use of *short supports* — by examining a subset of the variables, they can infer support (a justification that a variable-value pair may still form part of an assignment that satisfies the constraint) for all other variables and values and save substantial work – but short supports have not been studied in their own right. The two main contributions of this paper are the identification of short supports as important for constraint propagation, and the introduction of HAGGISGAC, an efficient and effective general purpose propagation algorithm for exploiting short supports. Given the complexity of HAGGISGAC, we present it as an optimised version of a simpler algorithm SHORTGAC. Although experiments demonstrate the efficiency of SHORTGAC compared with other general-purpose propagation algorithms where a compact set of short supports is available, we show theoretically and experimentally that HAGGISGAC is even better. We also find that HAGGISGAC performs better than GAC-Schema on full-length supports. We also introduce a variant algorithm HAGGISGAC-STABLE, which is adapted to avoid work on backtracking and in some cases can be faster and have significant reductions in memory use. All the proposed algorithms are excellent for propagating disjunctions of constraints. In all experiments with disjunctions we found our algorithms to be faster than Constructive Or and GAC-Schema by at least an order of magnitude, and up to three orders of magnitude.

## 1. Introduction

Constraint solvers typically employ a systematic backtracking search, interleaving the choice of an assignment of a decision variable with the *propagation* of the constraints to determine the consequences of the assignment made. Propagation algorithms can broadly be divided into two types. The first are specialised to reason very efficiently about constraint patterns that occur frequently in models. Examples include the global cardinality constraint (Régin, 1996) and the element constraint (Gent, Jefferson, & Miguel, 2006b). It is not feasible to support every possible constraint expression with a specialised propagator in this way, in which case general-purpose constraint propagators, such as GAC-Schema (Bessière & Régin, 1997), GAC2001/3.1 (Bessière, Régin, Yap, & Zhang, 2005), STR2 (Lecoutre, 2011) or MDDC (Cheng & Yap, 2010) are used. These are typically more expensive than specialised propagators but are an important tool when no specialised propagator is available.

A *support* in a constraint for a domain value of a variable is a justification that the value may still form part of an assignment that satisfies the constraint. It is usually given in terms of a set of *literals*: variable-value pairs corresponding to possible assignments to the other variables in the constraint. One of the efficiencies typically found in specialised propagators is the use of *short supports*: by examining a subset of the variables, they can infer support for all other variables and values and save substantial work. This use is typically implicit, i.e. achieved through a specialised algorithm which does not examine all variables in all cases. One of our contributions is to highlight the general importance of short supports.

As an example, consider the element constraint $x_y = z$, with $x_0, x_1, x_2, y \in \{0 \dots 2\}$, $z \in \{0 \dots 3\}$. This constraint is satisfied iff the element in position $y$ of vector $[x_0, x_1, x_2]$ equals $z$. Consider the set of literals $S = \{x_0 \mapsto 1, y \mapsto 0, z \mapsto 1\}$. This set clearly satisfies the definition of the constraint $x_y = z$, but it does not contain a literal for each variable. Any extension of $S$ with valid literals for variables $x_1$ and $x_2$ is a support. $S$ is an example of a *short support*.

In our previous work we introduced ShortGAC (Nightingale, Gent, Jefferson, & Miguel, 2011), a general-purpose propagation algorithm that exploits short supports. Until the introduction of ShortGAC, general-purpose propagators relied upon supports involving all variables. In this paper we develop the concept further and introduce a new algorithm HaggisGAC,[1] which is consistently more efficient than ShortGAC. Where available, the use of compact sets of short supports allows HaggisGAC to outperform greatly existing general-purpose propagation algorithms. In some cases, HaggisGAC even approaches the performance of special-purpose propagators. HaggisGAC is also very well suited to propagating disjunctions of constraints, and outperforms the traditional Constructive Or algorithm (Lagerkvist & Schulte, 2009; Würtz & Müller, 1996) by orders of magnitude. HaggisGAC is also more efficient than GAC-Schema on full-length supports. We also describe a variant, HaggisGAC-Stable, in which supports do not need to be deleted on backtracking. Applied to full-length supports, this version has greatly reduced memory usage.

ShortGAC, HaggisGAC and HaggisGAC-Stable are all instantiated with a function named findNewSupport (and are similar to GAC-Schema in this way). This function can be specific to a constraint, and generate short supports procedurally. Alternatively, a generic findNewSupport can retrieve short supports from a data structure.

Section 2 presents the necessary background, and Section 3 introduces the concept of short support. Section 4 outlines the basic idea used to deal with implicit supports throughout the paper. Section 5 gives full details of ShortGAC, including the complexity of key operations and alternative implementations for when short supports are provided in list form. Section 6 presents the new algorithm HaggisGAC as a development of ShortGAC. Both ShortGAC and HaggisGAC are evaluated experimentally in Section 7. Section 8 describes HaggisGAC-Stable, with corresponding experiments in Section 9. Finally, Sections 10 and 11 discuss related work and present our conclusions.

---

1. HaggisGAC is named for the legendary wild haggis of Scotland, which has both short legs and long legs for walking around hills. Like its namesake, HaggisGAC copes with both full-length and shorter supports and originates in Scotland. Details of the wild haggis can be found on Wikipedia, `http://en.wikipedia.org/wiki/Wild_haggis`, and in the *Veterinary Record* (King, Cromarty, Paterson, & Boyd, 2007).

## 2. Supports, GAC, Triggers

A *constraint satisfaction problem* (CSP) is defined as a set of variables $X$, a function that maps each variable to its domain, $D : X \to 2^{\mathbb{Z}}$ where each domain is a finite set, and a set of constraints $C$. A constraint $c \in C$ is a relation over a subset of the variables $X$. The *scope* of a constraint $c$, named scope($c$), is the set of variables that $c$ constrains.

A *solution* to a CSP is a function $s : X \to \mathbb{Z}$ that maps each variable $x \in X$ to a value from $D(x)$, such that for every constraint $c \in C$, the values of scope($c$) form a tuple that is in $c$ (i.e. the constraint is *satisfied*).

During a systematic search for a solution to a CSP, values are progressively removed from the domains $D$. Therefore, we distinguish between the *initial* domains and the *current* domains. The function $D$ refers to the current domains unless stated otherwise. A *literal* is defined as a variable-value pair, and is written $x \mapsto v$. A literal $x \mapsto v$ is *valid* if $v$ is in the current domain of $x$ (i.e. $v \in D(x)$).

**Definition 2.1.** *[Support] A **support** $S$ for constraint $c$ and domains $D$ is defined as a set of valid literals that contains exactly one valid literal for each variable in scope($c$) and satisfies $c$. Where necessary for disambiguation, we call such a support a **full-length support** or simply **long support**, to contrast with short supports as defined later.*

A property commonly established by constraint propagation algorithms is *generalised arc consistency* (GAC) (Mackworth, 1977). A constraint $c$ is GAC if and only if there exists a full-length support for every valid literal of every variable in scope($c$). GAC is established by identifying all literals $x \mapsto v$ for which no full-length support exists and removing $v$ from the domain of $x$. We consider only algorithms for establishing GAC in this paper.

A GAC propagation algorithm is usually situated in a systematic search. Hence, it must operate in three contexts: initialisation (at the root node), where support is established from scratch; following the deletion of one or more domain values (as a result of a branching decision and/or the propagation of other constraints), where support must be re-established selectively; and upon backtracking, where data structures must be restored to the correct state for this point in search. Our primary focus will be on the second context, operation following value deletion, although we will discuss efficient backtracking in Section 8. A GAC propagation algorithm would typically be called for each deleted domain value in turn. Once the algorithm has been called for each such domain value, the constraint will be GAC.

The propagation algorithms we present have the concept of *active support*, inspired by GAC-Schema (Bessière & Régin, 1997). An *active support* is a support that is currently in use to support a set of literals. Each literal has a set of active supports that support it. When an active support is found to be invalid, it is removed. When the set for some literal is empty, we say the literal has *lost support*. A new support is sought for the literal, and if found the new support becomes active. If no new support is found, the literal has no support and it is deleted.

In the propagation algorithms we present, for efficiency we make use of 'watched literals' as provided in Minion (Gent et al., 2006b), because propagators need not be called for *every* deleted domain value to establish GAC. We say that propagators attach and remove *triggers* on literals. When a domain value $v$ for variable $x$ is deleted, the propagator is called if and only if it has a trigger attached to the literal $x \mapsto v$. Doing so means that when a literal is deleted which is not attached to a trigger, *zero* work is incurred. We should emphasise that the use of watched literals is not fundamental to our work. If they are not available in a given solver, our algorithms only need a minor adaptation. When called on any literal removal, we may just return immediately if the literal is not in any active support, which can be checked in time $O(1)$. Thus our algorithms fit the traditional fine-grained scheme (Bessière & Régin, 1997) *except* that in some cases they will not be invoked because they use watched literals.

## 3. Short Supports

The concept of a *short support* is a generalisation of full-length support. It is defined below.

**Definition 3.1.** *[Short support] A **short support** $S$ for constraint $c$ and domains $D$ is defined as a set of valid literals $x \mapsto v$ such that $x \in \text{scope}(c)$, $x$ occurs only once in $S$, and every superset of $S$ that contains one valid literal for each variable in $\text{scope}(c)$ is a full-length support. A **strict short support** is a short support that is not a full-length support.*

The definition of short support includes both extremes. The empty set is a short support when the constraint is entailed (i.e. every tuple on $\text{scope}(c)$ within $D$ satisfies the constraint). Similarly, every full-length support $S$ is necessarily a short support, because the only superset of $S$ is itself. In our case studies we will see examples of both empty short supports and short supports that also happen to be full length.

Short supports can be used to maintain GAC. Just as with a full-length support, a short support provides GAC support for each literal contained within it. We call this *explicit support* for those literals. The new feature is that a short support also provides support for all valid literals of all variables not contained in the short support. This is because, by definition, every valid extension of the short support to cover all variables in $\text{scope}(c)$ is a full-length support. We say that a short support gives *implicit* GAC support for all valid literals of variables not in the short support.

We also define the concept of a complete set of short supports for a constraint.

**Definition 3.2.** *[Short support set] A **short support set** $\mathbb{S}(c, D)$ is a set of short supports for constraint $c$ under domains $D$, such that every full-length support $S$ of $c$ under $D$ is a (not necessarily strict) superset of at least one short support $S' \in \mathbb{S}(c, D)$.*

A constraint may have many short support sets. This gives us some latitude to implement one that is efficient to compute.

It is natural to ask how we can identify correct short supports given a constraint $c$. A simple but fundamental result is given in Lemma 3.3.

**Lemma 3.3.** *Given a constraint $c$ and domains $D$, the empty set $\{\}$ is a short support for $c$ iff GAC propagation for the constraint $not(c)$ leads to an empty domain.*

*Proof.* $\{\}$ is a short support if and only if every valid assignment to variables in $\text{scope}(c)$ satisfies $c$. Every assignment satisfies $c$ iff every assignment violates $not(c)$. If every assignment violates $not(c)$, then GAC propagation for the constraint $not(c)$ leads to an empty domain. To complete this last equivalence, note that if any assignment does not violate $not(c)$, all literals in that assignment are supported, so GAC propagation cannot cause an empty domain. □

This lemma has two important consequences. First, we can check *any* short support for correctness, not just the empty support. To check a short support $S = \{x_1 \mapsto v_1, \ldots, x_k \mapsto v_k\}$, we can simply set $D(x_1) = \{v_1\}, \ldots, D(x_k) = \{v_k\}$. All assignments now extend $S$, so $S$ is a short support iff $\{\}$ is. Lemma 3.3 applies so we can check the correctness of $S$ by propagating $not(c)$ and seeing if a domain is emptied.

The second consequence is negative, however. Determining whether GAC propagation will empty a domain is polynomially equivalent to actually performing GAC propagation (Bessière, Hebrard, Hnich, & Walsh, 2007). Since some constraints are NP hard to GAC propagate, it follows that it is not easy even to check if the empty set is a short support. Thus we cannot expect to find a method which is both fast and general for finding short supports for a constraint.

Given the provable difficulty of finding short supports from a set of full-length supports, we construct sets of short supports specifically for each of three experimental case studies in Section 7. The focus of this paper is to show the value of strict short supports if they are given to the system. The situation is analogous with that in an important area of constraints, namely that of exploiting symmetries in constraint problems (Gent, Petrie, & Puget, 2006). A large majority of research has assumed that sets of symmetries are provided to the system, even though finding such sets is hard. This has not inhibited research in exploiting symmetry, within which the automated detection of symmetry has become an important subarea (Mears, 2009; Puget, 2005): however we leave the automated construction of compact short support sets to future research. Analogously to patterns such as matrix symmetries (Flener, Frisch, Hnich, Kiziltan, Miguel, Pearson, & Walsh, 2002), we can at least identify a pattern which often lets us identify strict short supports, as we now describe.

## 3.1 Short Supports and Disjunction

Strict short supports arise naturally from disjunctions. If a constraint can be expressed as a disjunction of shorter constraints, then a set of strict short supports can be constructed for it. Suppose we have the following constraint.

$$c(x_1, x_2, x_3, x_4) \equiv c_1(x_1, x_2) \vee c_2(x_2, x_3) \vee c_3(c_3, x_4)$$

Suppose also that $A = \{x_1 \mapsto 2, x_2 \mapsto 1\}$ is a valid assignment that satisfies $c_1$. If we satisfy $c_1$, we satisfy $c$ regardless of the values of $x_3$ and $x_4$. Therefore $A = \{x_1 \mapsto 2, x_2 \mapsto 1\}$ is a strict short support for $c$.

**Lemma 3.4.** *Given constraint $c$, a domain set $D$, and a set of constraints $\{c_1 \ldots c_k\}$ where $\forall c_i \in \{c_1 \ldots c_k\} : scope(c_i) \subseteq scope(c)$ and $c \equiv c_1 \vee \cdots \vee c_k$, the following is a short support set (where we write $\mathrm{fls}(c_i, D)$ to mean the full-length supports of $c_i$ w.r.t. domains $D$):*

$$\mathbb{S}(c, D) = \{S \mid S \in \mathrm{fls}(c_1, D) \vee \cdots \vee S \in \mathrm{fls}(c_k, D)\}$$

*Proof.* (a) Each element of $\mathbb{S}(c, D)$ is a short support according to Definition 3.1 by the semantics of disjunction. (b) $\mathbb{S}(c, D)$ is a short support set by Definition 3.2. Every full-length support of $c$ must satisfy some disjunct $c_i$, therefore the full-length support contains a full-length support for $c_i$ that is included in $\mathbb{S}(c, D)$. □

Lemma 3.4 allows a short support set to be created for any disjunction, given the initial domains. We do this for two of our three case studies (for the third, the set is prohibitively large).

Using a similar approach to Lemma 3.4 we can create a function that generates short supports on demand. The function takes a valid literal $x \mapsto v$ and the current domains $D$, and returns a short support that supports $x \mapsto v$ (explicitly or implicitly), or NULL if none exists. The function can be constructed as follows. We create new domains $D'$ where $D'(x) = \{v\}$, and otherwise $D'$ is identical to $D$. If no disjunct is satisfiable under $D'$, then the function returns NULL. Otherwise, the function picks any disjunct $c_k$ that is satisfiable under $D'$, and returns a satisfying assignment of $c_k$ that is valid under $D'$. For each of the three case studies in Section 7, we created a function that follows this scheme with some optimisations.

Propagating disjunctions is recognised to be an important topic. Many papers have been published in this area (Würtz & Müller, 1996; Lhomme, 2003; Lagerkvist & Schulte, 2009; Jefferson, Moore, Nightingale, & Petrie, 2010). Exploiting strict short supports in the algorithms SHORTGAC, HAGGISGAC and HAGGISGAC-STABLE allows us to outperform the traditional Constructive Or algorithm (Würtz & Müller, 1996) by orders of magnitude.

## 3.2 Backtrack Stability of Short Supports

Within a search tree, propagation algorithms often spend significant time backtracking data structures. Reducing or eliminating backtracking can improve efficiency. For example, avoiding backtracking triggers can speed up a simple table propagator by more than 2 times (Gent et al., 2006b), and MAC-6 and MAC-7 can be much more efficient (in both space and time) if backtracking is avoided (Régin, 2005). There are two potential advantages of reducing use of backtracking state: it saves time restoring data structures, and it saves space by avoiding storing supports on the backtrack stack.

**Definition 3.5.** *[Backtrack Stable] A short support of constraint $c$ with current domains $D$ is backtrack stable iff it always remains a short support (according to Definition 3.1) after backtracking up the search tree.*

A short support $s$ may support some variable $x$ implicitly, and as we backtrack we may add values back into the domain of $x$ that are *not* consistent with $s$, meaning that $s$ no longer meets the definition of a short support. We give an example below.

**Example 3.1.** *Consider the constraint $b \to M[x] = y$, for a boolean variable $b$, array of variables $M$ and variables $x$ and $y$. When $b$ is assigned FALSE, this constraint is entailed, and so the empty short support can be used to support all literals in $M, x$ and $y$. This support is not backtrack stable, as on backtracking when TRUE is restored to the domain of $b$, the empty set is no longer a short support.*

Any support that is full length is backtrack stable: whenever the support is valid it supports all literals it contains. Backtrack stable supports always exist because we can use full-length supports in all cases (as in GAC-Schema), although these may be much longer than necessary.

In Section 8 we exploit backtrack stability to define a new algorithm.

## 4. SHORTGAC: An Overview

This section summarises the key ideas of the SHORTGAC propagation algorithm, along with an illustrative example.[2]

SHORTGAC maintains a set of short supports sufficient to support all valid literals of the variables in the scope of the constraint it is propagating. We refer to these as the *active* supports. The algorithm rests on exploiting the observation that, using short supports, support can be established for a literal in two ways. First, as usual, a short support that contains a literal supports that literal. Second, a literal $x \mapsto v$ is supported by a short support that contains no literal of variable $x$. Hence, the only short supports that do *not* support $x \mapsto v$ are those which contain a literal $x \mapsto w$ for some other value $w \neq v$.

The following data structures are central to the operation of the SHORTGAC algorithm:

**numSupports** is the total number of active short supports.

**supportsPerVar** is an array (indexed by $[x]$) indicating the number of active short supports containing each variable $x$.

**supportListPerLit** is an array (indexed by $[x \mapsto v]$) of lists of active short supports containing each literal $x \mapsto v$.

If the number of supports containing some variable $x$ is less than the total number of supports then there exists a support $s$ that does not contain $x$. Therefore, $s$ supports all literals of $x$. The algorithm spends no time processing variables all of whose literals are known to be supported in this way. Only for variables involved in all active supports do we have to seek support for literals with no active supports.

To illustrate, we consider the element example from the introduction: $x_y = z$, with $x_0, x_1, x_2, y \in \{0 \ldots 2\}$, and $z \in \{0 \ldots 3\}$. This constraint is satisfied iff the element in position $y$ of vector $[x_0, x_1, x_2]$ equals $z$. Suppose in the current state SHORTGAC is storing just one support: $A = \{x_0 \mapsto 1, y \mapsto 0, z \mapsto 1\}$. The data structures are as follows, where $\times$ indicates that a literal is not valid.[3]

---

2. The details we present here are different from those we presented previously (Nightingale et al., 2011), as we have optimised the data structures and algorithms compared with our previous work. The two most significant changes are: we no longer keep a count of supports per literal, saving overhead in maintaining this; and data is stored in a one dimensional vector by literal, instead of a two dimensional array by variable/value, saving space if variables in a constraint have very different domain sizes. Experiments in Appendix A demonstrate that the algorithms and data structures presented here perform better than our previous implementation.

3. For clarity, we have presented the one-dimensional array **supportListPerLit** in a two-dimensional format.

| Supports: | $A$: | | $x_0 \mapsto 1, y \mapsto 0, z \mapsto 1$ | | |
|---:|:---:|:---:|:---:|:---:|:---:|
| **supportListPerLit:** | Variable | | | | |
| Value | $x_0$ | $x_1$ | $x_2$ | $y$ | $z$ |
| 0 | {} | {} | {} | {$A$} | {} |
| 1 | {$A$} | {} | {} | {} | {$A$} |
| 2 | {} | {} | {} | {} | {} |
| 3 | × | × | × | × | {} |
| **supportsPerVar:** | 1 | 0 | 0 | 1 | 1 |
| **numSupports:** | 1 | | | | |

All values of $x_1$ and $x_2$ have support, since their **supportsPerVar** counters are both less than **numSupports**. Therefore the SHORTGAC algorithm can ignore $x_1$ and $x_2$ and only look for new supports of $x_0$, $y$ and $z$. Consider finding a new support for literals in $z$. SHORTGAC can ignore the literals with at least one support – in this case $z \mapsto 1$. The algorithm looks for literals $z \mapsto a$ where **supportListPerLit**$[z, a] = \{\}$. Here, $z \mapsto 0$ is such a literal, so SHORTGAC seeks a new support for it. A possible new support is $B = \{x_1 \mapsto 0, y \mapsto 1, z \mapsto 0\}$. Following its discovery, we update the data structures:

| Supports: | $A$: | | $x_0 \mapsto 1, y \mapsto 0, z \mapsto 1$ | | |
|---:|:---:|:---:|:---:|:---:|:---:|
| | $B$: | | $x_1 \mapsto 0, y \mapsto 1, z \mapsto 0$ | | |
| **supportListPerLit:** | Variable | | | | |
| Value | $x_0$ | $x_1$ | $x_2$ | $y$ | $z$ |
| 0 | {} | {$B$} | {} | {$A$} | {$B$} |
| 1 | {$A$} | {} | {} | {$B$} | {$A$} |
| 2 | {} | {} | {} | {} | {} |
| 3 | × | × | × | × | {} |
| **supportsPerVar:** | 1 | 1 | 0 | 2 | 2 |
| **numSupports:** | 2 | | | | |

Now variable $x_0$ is also fully supported, since **supportsPerVar**$[x_0] <$ **numSupports**. There remain three literals for which support has not been established: $y \mapsto 2, z \mapsto 2$ and $z \mapsto 3$. For the first two SHORTGAC finds supports such as $C = \{x_0 \mapsto 2, y \mapsto 0, z \mapsto 2\}$ and $D = \{x_2 \mapsto 0, y \mapsto 2, z \mapsto 0\}$. No support exists for $z \mapsto 3$, so 3 will be deleted, giving:

| Supports: | $A$: | | $x_0 \mapsto 1, y \mapsto 0, z \mapsto 1$ | | |
|---:|:---:|:---:|:---:|:---:|:---:|
| | $B$: | | $x_1 \mapsto 0, y \mapsto 1, z \mapsto 0$ | | |
| | $C$: | | $x_0 \mapsto 2, y \mapsto 0, z \mapsto 2$ | | |
| | $D$: | | $x_2 \mapsto 0, y \mapsto 2, z \mapsto 0$ | | |
| **supportListPerLit:** | Variable | | | | |
| Value | $x_0$ | $x_1$ | $x_2$ | $y$ | $z$ |
| 0 | {} | {$B$} | {$D$} | {$A, C$} | {$B, D$} |
| 1 | {$A$} | {} | {} | {$B$} | {$A$} |
| 2 | {$C$} | {} | {} | {$D$} | {$C$} |
| 3 | × | × | × | × | × |
| **supportsPerVar:** | 2 | 1 | 1 | 4 | 4 |
| **numSupports:** | 4 | | | | |

All valid literals are now supported. Nothing further need be done until a change in state, such as the removal of a value by a branching decision or propagation.

## 5. ShortGAC: **Details**

The key tasks in implementing ShortGAC are: data structure update; iteration over variables where supportsPerVar equals numSupports; and iteration over the unsupported values of a variable. This section describes the infrastructure that allows us to perform each of these tasks efficiently.

### 5.1 ShortGAC **Data Structures**

An active short support $S$ of arity $k$ provides explicit support for each of the $k$ literals it contains. Therefore, a reference to $S$ must appear in $k$ of the lists of supportListPerLit. To do this, we represent $S$ with two types of object: ShortSupport and ShortSupportCell. The ShortSupport object contains $k$ ShortSupportCell objects, each of which contains a literal $x \mapsto v$,[4] and a reference to the parent ShortSupport. The elements of the array support-ListPerLit are doubly-linked lists of ShortSupportCells. Through the reference to the parent ShortSupport, we can iterate through all active short supports for a given literal.

The algorithm iterates over all variables $x$ where supportsPerVar$[x]$ equals numSupports. The following data structure represents a partition of the variables by the number of supports. It allows constant time size checking and linear-time iteration of each cell in the partition, and allows any variable to be moved into an adjacent cell (i.e. if the number of supports increases or decreases by 1) in constant time. It is inspired by the indexed dependency array in Gecode (Schulte & Tack, 2010).

**varsBySupport** is an array containing a permutation of the variables. Variables are ordered by non-decreasing number of active supports (supportsPerVar$[x]$).

**supportNumLowIdx** is an array of integers, indexed from 0 to the number of literals, that being the maximum number of active supports possible. Either supportNumLowIdx$[i]$ is the smallest index in varsBySupport with $i$ or more active supports, or (when there are no such variables) supportNumLowIdx$[i]= k$ where $k$ is the total number of variables. $k$ acts as a sentinel value. The set of variables with $i$ supports is:

$$\text{varsBySupport}[\text{supportNumLowIdx}[i] \ldots \text{supportNumLowIdx}[i+1] - 1]$$

Initially, all variables have 0 active supports, so supportNumLowIdx$[0] = 0$ and the rest of the array is set to $k$.

The following table illustrates how the partition data structure works (on a different example with 11 variables). Suppose supportsPerVar$[x_2]$ changed from 7 to 6. $x_2$ and $y_1$ (boxed) are swapped in varsBySupport and the cell boundary is moved so that $x_2$ is in the lower cell. Consequently, supportNumLowIdx$[7]$ is incremented by 1.

| varsBySupport[] | $w_1$ | $w_2$ | $y_1$ | $x_1$ | $x_2$ | $y_2$ | $y_3$ | $x_3$ | $z_1$ | $z_2$ | $z_3$ |
|---|---|---|---|---|---|---|---|---|---|---|---|
| supportsPerVar | 6 | 6 | 7 | 7 | 7 | 7 | 7 | 7 | 8 | 8 | 8 |
| $x_2$ updated | $w_1$ | $w_2$ | $\boxed{x_2}$ | $x_1$ | $\boxed{y_1}$ | $y_2$ | $y_3$ | $x_3$ | $z_1$ | $z_2$ | $z_3$ |

---

4. A literal $x \mapsto v$ is represented using a single integer $i$. There is a mapping between $x \mapsto v$ and $i$, which allows O(1) access to $x$ and $v$ from $i$ and vice-versa.

**Require:** *sup*: a ShortSupport
 1: **for all** *sc*: ShortSupportCell $\in$ *sup* **do**
 2:   $(x \mapsto v) \leftarrow sc.literal$
 3:   **if** supportListPerLit$[x \mapsto v] = \{\}$ **then**
 4:     attachTrigger$(x \mapsto v)$
 5:   Add *sc* to doubly linked list supportListPerLit$[x \mapsto v]$
 6:   supportsPerVar$[x]$++
 7:   $sx \leftarrow$ supportsPerVar$[x]$
 8:   *cellend* $\leftarrow$ supportNumLowIdx$[sx]-1$
 9:   swap$(x,$ varsBySupport$[cellend])$
10:   supportNumLowIdx$[sx]$--
11: numSupports++

Procedure 1: addSupport($sup$)

For a variable $x$ with supportsPerVar$[x] =$ numSupports, SHORTGAC iterates over the values with zero explicit supports. To avoid iterating over all values, we use a set data structure:

**zeroLits** is an array (indexed by $[x]$) of stacks containing the literals of variable $x$ with zero explicit support, in no particular order.

**inZeroLits** is an array (indexed by $[x \mapsto v]$) of booleans indicating whether literal $x \mapsto v \in$ zeroLits$[x]$.

When supportListPerLit$[x \mapsto v]$ is reduced to the empty list, if inZeroLits$[x \mapsto v]$ is false then $x \mapsto v$ is pushed onto zeroLits$[x]$ (and inZeroLits$[x \mapsto v]$ is set to true). As an optimisation, values are not eagerly removed from the set; they are only removed lazily when the set is iterated. Also, the set is not backtracked. During iteration, a non-zero value is removed by swapping it to the top of the stack, and popping. This lazy maintenance never costs work overall because, if the value would have been removed eagerly, then it will be removed the next time the set is iterated, costing $O(1)$. It can save work, because we may never iterate over the list before the value would have been restored to the set again.

We use a free list to manage the set of **ShortSupport** objects to avoid the cost of unnecessary object construction/destruction. The **ShortSupport** object retrieved from the free list may contain too few **ShortSupportCell** objects, so we use a resizable vector data structure. The size is only ever increased.

## 5.2 Adding and Deleting Supports

When a support is added or deleted, all the data structures described above must be updated. This is done by Procedures 1 (addSupport) and 2 (deleteSupport). Both these procedures iterate through the given short support, and for each literal in it they update supportListPerLit, supportsPerVar, varsBySupport and supportNumLowIdx. Procedure 2 also inserts the literal into zeroLits if necessary. We briefly explain the maintenance of varsBySupport as it will become important in Section 6.2. Suppose we are adding support for literal $x \mapsto v$ in Procedure 1. Because it has an additional support, $x$ must be moved to the next cell in varsBySupport. Line 8 finds the end of the cell that $x$ is in, then we swap $x$ to the

**Require:** $sup$: a ShortSupport
1: **for all** $sc$: ShortSupportCell $\in sup$ **do**
2:    $(x \mapsto v) \leftarrow sc.literal$
3:    Remove $sc$ from doubly-linked list supportListPerLit$[x \mapsto v]$
4:    supportsPerVar$[x]$--
5:    **if** supportListPerLit$[x \mapsto v] = \{\}$ **then**
6:      removeTrigger$(x \mapsto v)$
7:      **if** $\neg$inZeroLits$[x \mapsto v]$ **then**
8:        inZeroLits$[x \mapsto v] \leftarrow true$
9:        zeroLits$[x]$.push$(x \mapsto v)$
10:   $sx \leftarrow$ supportsPerVar$[x]$
11:   $cellend \leftarrow$ supportNumLowIdx$[sx+1]$
12:   swap$(x,$ varsBySupport$[cellend])$
13:   supportNumLowIdx$[sx+1]$++
14: numSupports--

<div align="center">Procedure 2: deleteSupport($sup$)</div>

**Require:** $x \not\mapsto v$ (where $v$ has been pruned from the domain of $x$)
1: **while** supportListPerLit$[x \mapsto v] \neq \{\}$ **do**
2:   deleteSupport(supportListPerLit$[x \mapsto v]$.pop())
3: **repeat**
4:   continueLoop $\leftarrow$ false
5:   **for all** $i \in \{$supportNumLowIdx[numSupports]$\ldots$ supportNumLowIdx[numSupports+1]-1$\}$ **do**
6:     $y \leftarrow$ varsBySupport$[i]$
7:     **if** SHORTGAC-variableUpdate$(y) =$ true **then**
8:       continueLoop $\leftarrow$ true
9:       **break** out of for loop Line 5
10: **until** continueLoop = false

<div align="center">Procedure 3: SHORTGAC-Propagate: propagate($x \not\mapsto v$)</div>

end of its cell using a subroutine swap($x_i$, $x_j$). This simple procedure (not given) locates and swaps the two variables in varsBySupport, leaving other variables unaffected. To do so it makes use of a second array, varsBySupInv, which is the inverse mapping of varsBySupport. Having done this, the cell boundary is decremented so that (in its new position), $x$ is now in the higher cell. Another point to note is that addSupport will add a trigger for $x \mapsto v$ if $sup$ is the only active explicit support to contain that literal, while deleteSupport will remove the trigger if the deleted support is the only support.

Finally, we note that we do not have special-purpose methods to undo these changes on backtracking. On backtracking past the point where a support is added, we simply call deleteSupport, and similarly we call addSupport when we backtrack past a support's deletion.

**Require:** variable $x$
1: **for all** $(x \mapsto v) \in \mathsf{zeroLits}[x]$ **do**
2:     **if** $\mathsf{supportListPerLit}[x \mapsto v] \neq \{\}$ **then**
3:         Remove $(x \mapsto v)$ from $\mathsf{zeroLits}[x]$
4:     **else**
5:         **if** $v \in D(x)$ **then**
6:             sup $\leftarrow$ findNewSupport$(x \mapsto v)$
7:             **if** sup = Null **then**
8:                 prune$(x \mapsto v)$
9:             **else**
10:                addSupport(sup)
11:                **if** $\mathsf{supportListPerLit}[x \mapsto v] \neq \{\}$ **then**
12:                    Remove $(x \mapsto v)$ from $\mathsf{zeroLits}[x]$
13:                **return** true
14: **return** false

Procedure 4: ShortGAC-variableUpdate: $(x)$. Here and in other pseudocode we abstract the detailed maintenance of the $\mathsf{zeroLits}$ and $\mathsf{inZeroLits}$ data structures. It might seem that the test on Line 11 must always succeed. However, although *sup* must *support* $x \mapsto v$, it does not have to *contain* $x \mapsto v$ as it might be an implicit support. The findNewSupport function is discussed in Section 5.5.

## 5.3 The Propagation Algorithm

The ShortGAC propagator (Procedure 3) is only invoked when a literal contained in one or more active short supports is pruned.[5] It first deletes all supports involving the pruned literal. Then it checks all variables $y$ which are not implicitly supported, i.e. where $\mathsf{supportsPerVar}[y]=\mathsf{numSupports}$ (Line 5). Each such variable $y$ is checked by Procedure 4 (ShortGAC-variableUpdate, described below). If this call results in a new support being found, then the data structures will have changed (ShortGAC-variableUpdate$(y)$ returns true to indicate this) and we must break out of the for-all-loop (Line 9) and go round again. Iteration therefore continues until either no new support is necessary or no new support can be found.

ShortGAC-variableUpdate (Procedure 4) is used to check the status of every variable lacking implicit support. It iterates over $\mathsf{zeroLits}$, i.e. the literals for a variable which might have zero explicit supports. Since $\mathsf{zeroLits}$ is maintained lazily, on each iteration we first check that the literal does indeed have no explicit support, and correct $\mathsf{zeroLits}$ if necessary (Lines 2–3). The important case is that the literal indeed has no support. Then, provided that $v$ is in the current domain of $x$, we must seek a new support by calling findNewSupport for the constraint. If there is no support, value $v$ must be pruned from the domain of $x$, or if we have found a support we update data structures by calling addSupport.

To initialise data structures at the root of search, Lines 3–10 of Procedure 3 are invoked. Notice that these lines do not refer to the parameter $x \not\mapsto v$, and on first calling there are no supports at all so the initial iteration at Line 5 is over all variables.

---

5. As we noted earlier, if watched literals are not available in a solver, a simple check can be made at the start of the procedure, to return immediately if the removed literal is in no active support.

### 5.4 Complexity Analysis of SHORTGAC

In this section we provide a complexity analysis of SHORTGAC as it is used incrementally during search in a constraint solver. The analysis has as parameters the arity of the constraint $n$, the maximum domain size $d$, and the cost $f$ of calling findNewSupport. We assume that both attaching and removing a trigger to a literal are $O(1)$. This is the case in Minion 0.12.

First we observe that the swap procedure executes in $O(1)$ time: each operation in swap is $O(1)$ and it does not loop. Secondly we establish the time complexity of the procedures addSupport and deleteSupport, which are key to the algorithm.

**Lemma 5.1.** *Procedure 1 (addSupport) has time complexity $O(n)$.*

*Proof.* The outer loop on Line 1 iterates over the literals in the short support. In the worst case, there are $n$ literals. We now consider the steps within this loop. The list test on Line 3 is $O(1)$, as is the call to attachTrigger on Line 4. Adding the ShortSupportCell to the doubly-linked list on Line 5 is $O(1)$, as are the following five array dereferences. As established above, the swap procedure is also $O(1)$. Hence, addSupport is $O(n)$. □

**Lemma 5.2.** *Procedure 2 (deleteSupport) has time complexity $O(n)$.*

*Proof.* Similarly to the add Support procedure, the outer loop on Line 1 has at most $n$ iterations. The removal from the doubly-linked list on Line 3 is $O(1)$, as are the array dereferences on Line 4 and subsequently. The list test on Line 5 and the call to removeTrigger on Line 6 are both $O(1)$, as is the stack push operation on Line 9. Recalling once again that the swap procedure is $O(1)$, deleteSupport is $O(n)$. □

**Theorem 5.3.** *Procedure 3 (SHORTGAC-propagate) has time complexity in $O(n^2d^2+ndf)$. The upper bound can be obtained, i.e. the worst case time complexity is in $\Omega(n^2d^2+ndf)$.*

*Proof.* Analysis for the first statement breaks down into three parts.

First, the loop on Line 1 is over the elements of supportListPerLit. The worst case occurs when $nd$ literals have an explicit support. Of these supports, a maximum of $(n-1)d+1$ can involve a particular literal, because this literal may be in the short support for every literal of every other variable ($(n-1)d$), and itself (1). The cost of the body of this loop is $O(n)$ from Lemma 5.2, so the total is $O(n^2d)$. This will be dominated by the next part.

The second part is the loop from lines 3–10. The maximum number of iterations in Line 5 is $n$ when all supports are full length and so the iteration in Line 5 contains all $n$ variables. Successive calls to Procedure 4 at Line 7 can add at most $O(d)$ new supports. But each support addition triggers a restart of the loop beginning on Line 5 over all $n$ variables, for a total of at most $O(n^2d)$ calls to Procedure 4. Each such call involves $O(d)$ iterations of the loop on Line 1 of Procedure 4. Therefore the innermost loop is run at most $O(n^2d^2)$ times.

To complete the proof of the first statement, we consider the cost of the innermost loop of Procedure 4. Within this loop, most operations are $O(1)$, the exceptions being the call to findNewSupport on Line 6 (cost $f$) and the call to addSupport on Line 10 (cost $n$ from Lemma 5.1). But $f$ is the dominating cost, since it must at least traverse the new support to record it. However, of the $n^2d^2$ iterations, there can be at most $nd$ calls to findNewSupport, after which time all valid literals will have an explicit support. So the cost is either $O(n^2d^2)$ or $O(ndf)$, whichever is greater. In any case the cost is $O(n^2d^2 + ndf)$.

The upper bounds of $ndf$ and $n^2d^2$ can be attained in the worst case. If each literal needs a new support, we have $\Omega(ndf)$ calls to findNewSupport. We can have cost $\Omega(n^2d^2)$ if there are $nd$ literals with explicit support, each of size $n$, and each variable ends up with (for example) $d/2$ values supported and $d/2$ values deleted. The worst case is thus $\Omega(n^2d^2 + ndf)$. □

Procedure 3 can be invoked at most $n(d-1)$ times in one branch of the search tree, therefore the complexity for one branch is $O(n^3d^3 + n^2d^2f)$.

### 5.4.1 A SECOND COMPLEXITY ANALYSIS

The analysis above can be very conservative when the total number, and maximum size, of short supports is small. Therefore, we give another complexity analysis with two additional parameters: the maximum length $l$ of short supports returned by findNewSupport, and the total number $s$ of distinct short supports that may be returned by findNewSupport. This analysis also pertains to a branch of search rather than a single call to the propagate algorithm.

The first part of this complexity analysis concerns the $s$ short supports of length $l$. Each short support may be added to the active set once, and may be deleted once down a branch. Each short support must also be found by calling findNewSupport, with cost $O(f)$. Lemma 5.1 shows that the addSupport procedure takes $O(n)$ time. The same lemma can be re-stated in terms of $l$, because the loop in addSupport will iterate $O(l)$ times, giving a total time of $O(l)$. This also applies to deleteSupport. Since there are $s$ short supports, the cost of finding, adding and deleting (collectively *processing*) short supports is $O(s(l + f))$ down a branch.

Secondly, the algorithm may make calls to findNewSupport that return NULL. This can happen at most $n(d-1)+1$ times, because this is the maximum number of domain values that may be deleted. Therefore the cost is $O(ndf)$.

In addition, SHORTGAC does some operations that have not been charged to either of the above categories. To analyse these, we must do a top-down analysis of algorithm.

Procedure 3 is invoked $O(s)$ times (each time a short support is invalidated). Lines 1–2 are already charged to processing short supports. The body of the loop on lines 3–10 may be executed $s$ times when a new support is found, and a further $s$ times when no new support is found, therefore $O(s)$ times in total down a branch of search.

Now we come to the inner loop on lines 5–9. From Lemma 5.4 (below), unless a domain is empty there is always one or more active short support. Therefore, at most $l$ variables will be contained in all active short supports, so at most $l$ variables are in the relevant partition of varsBySupport, and the loop body will be executed $O(l)$ times.

**Lemma 5.4.** *After initialisation, Procedure 3 always has at least one active short support or a variable domain is empty.*

*Proof.* Suppose the opposite. The algorithm is invoked each time a literal in an active short support is pruned, therefore to delete all active short supports they must all contain one literal $x \mapsto v$. If all active short supports contain variable $x$, then all values in the domain of $x$ are not implicitly supported and must be explicitly supported. Therefore $v$ must be the last remaining value in $D(x)$. Now to prune $x \mapsto v$ empties the domain and we have a contradiction. □

Down a branch, this causes $O(sl)$ calls to ShortGAC-variableUpdate, on Line 7. Each call to ShortGAC-variableUpdate takes $O(d)$ time because there may be $d-1$ invalid literals or $d$ explicitly supported literals in zeroLits. Other time spent in this procedure is charged to processing short supports, or to pruning domains. Therefore in the top-down analysis the cost is $O(sld)$.

Overall, the time complexity is $O(s(l+f) + ndf + sld)$, a tighter bound in some cases than the one given in the section above. For example, a SAT clause has $s = n$, $f = n$, $l = 1$ and $d = 2$, giving a time complexity of $O(n^2)$ for a branch of search.

## 5.5 Instantiation of findNewSupport

Similarly to GAC-Schema (Bessière & Régin, 1997), ShortGAC must be instantiated with a findNewSupport function. The function takes a valid literal, and returns a support if one exists, otherwise returns Null. One way to do this is to write a specialist findNewSupport function for each constraint. We do this in each of the empirical case studies below. In each case, the findNewSupport function is much simpler than a propagator for the same constraint. We use Lemma 3.4 to build the findNewSupport functions, which reduces the task to finding satisfying tuples of simple constraints like $x < y$ and $x = y$.

The alternative is to write a generic version of findNewSupport for the case where all short supports are given as a list. We now detail two generic instantiations of findNewSupport for lists, and in our case studies below we compare them with the specialist functions.

### 5.5.1 findNewSupport-List

We provide a generic instantiation named findNewSupport-List (Procedure 5) that takes a list of short supports for each literal (supportList), including both the explicit and implicit short supports for that literal. This is analogous to the Positive instantiation of GAC-Schema (Bessière & Régin, 1997). FindNewSupport-List has persistent state: listPos, an array of integers indexed by variable and value, initially 0. This indicates the current position in the supportList. The algorithm simply iterates through the list of supports, seeking one where all literals are valid. ListPos is not backtracked, with the consequence that when the end of the list is reached, we cannot fail immediately and must search again from the start back to listPos. Down a branch of the search tree, any particular element of the list may be looked at more than once. However, this algorithm is optimal in both time and space across the search tree (Gent, 2012). This surprising result is achieved by amortizing the cost across all branches. Practically, using listPos stops the algorithm always starting from the first element of the list, and it seems to be a good tradeoff between avoiding provably unnecessary work and doing too much data structure maintenance.

A constraint-specific findNewSupport can sometimes find shorter supports than find-NewSupport-List. This is because a specific findNewSupport can take advantage of current domains whereas the supportList may only contain supports given the initial domains. For example, if the constraint becomes entailed, the specific findNewSupport can return the

**Require:** $x$, $v$, supportList
 1: **for all** $j \in \{\text{listPos}[x, v]\dots(\text{supportList}[x, v].\text{size-1})\}$ **do**
 2:     $sup \leftarrow \text{supportList}[x, v, j]$
 3:     **if** all literals in $sup$ are valid **then**
 4:         $\text{listPos}[x, v] \leftarrow j$
 5:         **return** $sup$
 6: **for all** $j \in \{0\dots\text{listPos}[x, v]{-}1\}$ **do**
 7:     $sup \leftarrow \text{supportList}[x, v, j]$
 8:     **if** all literals in $sup$ are valid **then**
 9:         $\text{listPos}[x, v] \leftarrow j$
10:         **return** $sup$
11: **return** NULL

Procedure 5: findNewSupport-List: findNewSupport($x$, $v$). The first block searches from the location of the previous support to the end of the support list. If it is unsuccessful the search restarts from the start of the list in the second block. This circular approach removes the need to backtrack listPos.

empty support whereas the list version we have presented cannot. We exploit this fact in Case Study 3 below.

### 5.5.2 FINDNEWSUPPORT-NDLIST

The list instantiation has two major disadvantages. First, it can be inefficient because it is unable to skip over sets of invalid tuples. The literature contains many solutions to this problem in the context of full-length supports, for example binary search (Lecoutre & Szymanek, 2006) or tries (Gent, Jefferson, Miguel, & Nightingale, 2007). Second, it can require a large amount of memory. For each short support $S$, there are potentially $nd$ pointers to $S$, because there is a pointer to it for each literal that $S$ implicitly supports.

In this section we give a second generic list instantiation based on NextDifference lists (Gent et al., 2007). We have a single list (named supportList) containing all short supports (indexed by an integer), and a second list named NDList where for each support $s =$ supportList$[j]$, for each literal in the support $s[k]$, NDList$[j][k]$ is the index of the next support that does *not* contain literal $s[k]$. Thus, when searching the list, the algorithm is able to jump over sets of short supports that all contain the same invalid literal. The version of findNewSupport for NextDifference lists is given in Procedure 6.

This approach solves both of the problems with the list instantiation: it is able to jump over sets of invalid short supports, and usually requires substantially less memory. In fact it it is optimal in space (unlike the list instantiation): given $t$ short supports of length at most $l$, the NextDifference list is $O(tl)$. However it uses only one list of supports, therefore it can spend time searching through short supports that do not support the desired literal.

### 5.6 Literals of Assigned Variables

Suppose SHORTGAC discovers a new support $S$ that contains a literal $x \mapsto v$, and $x$ is assigned to $v$. Since $x$ can take no value other than $v$, it is sound to remove $x \mapsto v$ from $S$ and save the overhead of adding it. We apply this minor optimisation in all cases when using SHORTGAC, and also in all cases when using HAGGISGAC (described in Section 6). How-

**Require:** $x$, $v$, supportList, NDList
1: $j \leftarrow$ listPos$[x, v]$
2: **while** $j <$ supportList.size **do**
3:     $sup \leftarrow$ supportList$[j]$
4:     $nextDiff \leftarrow$ NDList$[j]$
5:     **for** $k \in \{0 \ldots sup.\text{size} - 1\}$ **do**
6:         $(y \mapsto b) \leftarrow sup[k]$
7:         **if** $b \notin D(y)$ or $(x = y$ and $v \neq b)$ **then**
8:             $j \leftarrow nextDiff[k]$ {Jump to next short support where $y$ is assigned a different value.}
9:             **continue** while loop at Line 2
10:   listPos$[x, v] \leftarrow j$
11:   **return** $sup$
12: $j \leftarrow 0$
13: **while** $j <$ listPos$[x, v]$ **do**
14:     $sup \leftarrow$ supportList$[j]$
15:     $nextDiff \leftarrow$ NDList$[j]$
16:     **for** $k \in \{0 \ldots sup.\text{size} - 1\}$ **do**
17:         $(y \mapsto b) \leftarrow sup[k]$
18:         **if** $b \notin D(y)$ or $(x = y$ and $v \neq b)$ **then**
19:             $j \leftarrow nextDiff[k]$ {Jump to next short support where $y$ is assigned a different value.}
20:             **continue** while loop at Line 13
21:   listPos$[x, v] \leftarrow j$
22:   **return** $sup$
23: **return** NULL

Procedure 6: findNewSupport-NDlist: findNewSupport$(x, v)$

ever this optimisation cannot be used with HAGGISGAC-STABLE (described in Section 8) because that algorithm retains active supports as it backtracks, and after backtracking the literal $x \mapsto v$ may no longer be assigned.

## 6. HAGGISGAC: Dealing with Both Full-Length and Strict Short Supports

We now introduce HAGGISGAC. We show that it has better theoretical properties than SHORTGAC. Furthermore, experiments show it runs substantially faster in many cases on strict short supports than SHORTGAC (which is specialised for strict short supports), and substantially faster on full-length supports than GAC-Schema.

### 6.1 Introduction and Motivating Example

SHORTGAC is designed to exploit the concept of implicit support, but has some inefficiencies when dealing with explicit supports and especially full-length supports. Consider for example the constraint AllDifferentExceptZero, in which the constraint is that all non-zero values in the array must be different, but that zero may occur freely. This constraint might be used, for example, in a timetabling problem where classes taking place in different rooms must be different, but we use zero to represent a room being unused and this can occur multiple times. Suppose we have AllDifferentExceptZero$([w, x, y, z])$, each variable with initial domain $\{0, 1, 2, 3\}$. Supports for the constraint are full-length supports in which every non-zero value is different, *or* any three variables equalling zero where the last variable may take any value. Suppose we execute SHORTGAC and reach the following situation:

| Supports: | $A$: | | $w \mapsto 0, x \mapsto 2, y \mapsto 3, z \mapsto 1$ | |
|---|---|---|---|---|
| | $B$: | | $w \mapsto 0, x \mapsto 3, y \mapsto 2, z \mapsto 1$ | |
| | $C$: | | $w \mapsto 3, x \mapsto 0, y \mapsto 1, z \mapsto 2$ | |
| | $D$: | | $x \mapsto 0, y \mapsto 0, z \mapsto 0$ | |
| | $E$: | | $w \mapsto 0, x \mapsto 1, y \mapsto 2, z \mapsto 3$ | |
| supportListPerLit: | | | Variable | |
| Value | $w$ | $x$ | $y$ | $z$ |
| 0 | $\{A, B, E\}$ | $\{C, D\}$ | $\{D\}$ | $\{D\}$ |
| 1 | $\{\}$ | $\{E\}$ | $\{C\}$ | $\{A, B\}$ |
| 2 | $\{\}$ | $\{A\}$ | $\{B, E\}$ | $\{C\}$ |
| 3 | $\{C\}$ | $\{B\}$ | $\{A\}$ | $\{E\}$ |
| supportsPerVar: | 4 | 5 | 5 | 5 |
| numSupports: | | | 5 | |

Notice that the lack of explicit supports for $w \mapsto 1$ and $w \mapsto 2$ is acceptable because we have supportsPerVar$[w] = 4 <$ numSupports $= 5$. Now suppose the literal $y \mapsto 0$ is deleted by some other constraint. This causes support $D$ to be deleted, causing the following state:

| Supports: | $A$: | | $w \mapsto 0, x \mapsto 2, y \mapsto 3, z \mapsto 1$ | |
|---|---|---|---|---|
| | $B$: | | $w \mapsto 0, x \mapsto 3, y \mapsto 2, z \mapsto 1$ | |
| | $C$: | | $w \mapsto 3, x \mapsto 0, y \mapsto 1, z \mapsto 2$ | |
| | $E$: | | $w \mapsto 0, x \mapsto 1, y \mapsto 2, z \mapsto 3$ | |
| supportListPerLit: | | | Variable | |
| Value | $w$ | $x$ | $y$ | $z$ |
| 0 | $\{A, B, E\}$ | $\{C\}$ | $\times$ | $\{\}$ |
| 1 | $\{\}$ | $\{E\}$ | $\{C\}$ | $\{A, B\}$ |
| 2 | $\{\}$ | $\{A\}$ | $\{B, E\}$ | $\{C\}$ |
| 3 | $\{C\}$ | $\{B\}$ | $\{A\}$ | $\{E\}$ |
| supportsPerVar: | 4 | 4 | 4 | 4 |
| numSupports: | | | 4 | |

At this point SHORTGAC iterates through the zeroLits lists for all variables where supportsPerVar = numSupports, in this case all four variables. It will discover that we must find new supports for $w \mapsto 1, w \mapsto 2$ and $z \mapsto 0$. However, this is inefficient for two reasons. First, we should not need to check zeroLits$[z]$ to discover $z \mapsto 0$, because the support list for $z \mapsto 0$ became empty during the deletion of support $D$, so we could have discovered it then. Second, we should only need to look at zeroLits$[w]$ because the deletion of $D$ has caused $w$ to lose its implicit support. We should not need to check zeroLits for $x, y, z$ because these variables were not implicitly supported prior to $D$'s deletion. Removing these two reasons for inefficiency is the motivation behind our development of HAGGISGAC. In this example, it can focus directly on the literal $z \mapsto 0$ and the set zeroLits$[w]$ as the only literals potentially needing new support.

The fundamental problem with SHORTGAC is that it cannot efficiently detect when a literal loses its last support. Every variable with no implicit support is checked every time any support is deleted, so SHORTGAC can take $O(nd)$ time to find a single literal that needs a new support or to discover that there is no such literal. To improve upon this, we wish

| $i$ | 0 | 1 | 2 | 3 | 4 | 5 | 6 | 7 | 8 | 9 | 10 |
|---|---|---|---|---|---|---|---|---|---|---|---|
| varsBySupport$[i]$ | $w_1$ | $w_2$ | $y_1$ | $x_1$ | $x_2$ | $y_2$ | $y_3$ | $x_3$ | $z_1$ | $z_2$ | $z_3$ |
| supportsPerVar | 6 | 6 | 7 | 7 | 7 | 7 | 7 | 7 | 8 | 8 | 8 |
| $x_2$ updated | $w_1$ | $w_2$ | $x_2$ | $x_1$ | $y_1$ | $y_2$ | $y_3$ | $x_3$ | $z_1$ | $z_2$ | $z_3$ |
| $x_3$ updated | $w_1$ | $w_2$ | $x_2$ | $x_3$ | $y_1$ | $y_2$ | $y_3$ | $x_1$ | $z_1$ | $z_2$ | $z_3$ |
| $z_2$ updated | $w_1$ | $w_2$ | $x_2$ | $x_3$ | $y_1$ | $y_2$ | $y_3$ | $x_1$ | $z_2$ | $z_1$ | $z_3$ |
| $x_1$ updated | $w_1$ | $w_2$ | $x_2$ | $x_3$ | $x_1$ | $y_2$ | $y_3$ | $y_1$ | $z_2$ | $z_1$ | $z_3$ |
| $z_3$ updated | $w_1$ | $w_2$ | $x_2$ | $x_3$ | $x_1$ | $y_2$ | $y_3$ | $y_1$ | $z_2$ | $z_3$ | $z_1$ |
| $z_1$ updated | $w_1$ | $w_2$ | $x_2$ | $x_3$ | $x_1$ | $y_2$ | $y_3$ | $y_1$ | $z_2$ | $z_3$ | $z_1$ |
| supportsPerVar | 6 | 6 | 6 | 6 | 6 | 7 | 7 | 7 | 7 | 7 | 7 |

Figure 1: Illustration of how deleteSupport concentrates all variables that have just lost their last implicit support. See main text for the full description.

HaggisGAC to be able to detect the loss of a literal's last explicit support in time $O(1)$, and the loss of a variable's last implicit support in time $O(1)$. Perhaps surprisingly, both these goals are achievable by the use of data structures already existing in ShortGAC.

## 6.2 Finding Literals With No Support Efficiently

Of the two types of support, detecting when the last *explicit* support for a literal is lost is the simpler task. When we delete a support, Procedure 2 iterates through the literals in a short support. For each literal it removes a ShortSupportCell from the corresponding supportListPerLit and updates data structures appropriately. If the list is empty – tested at Line 5 of Procedure 2 – the literal has lost its last explicit support. We now add this literal to a scratch list of literals which have lost their last explicit support: we describe below how we process the scratch list. The additional cost is $O(1)$ when we detect an empty list. Because we are inside an existing test, there is zero additional cost when the literal has not lost its last support. This contrasts with ShortGAC which tests (in Procedure 4) every variable with no implicit support, for a worst case cost of $O(n)$ even when no literal has lost its last explicit support.

The more subtle task is to detect when a variable (and thus all literals involving it) has lost its last *implicit* support. The reason this is more difficult is that we are seeking variables that are *not* involved in the support being deleted, but in Procedure 2 we iterate through the literals that *are* in the support being deleted. The variables we seek are those $x$ which have supportsPerVar$[x] =$ numSupports after the support deletion, while they had supportsPerVar$[x] <$ numSupports before the support deletion. (Variables that have supportsPerVar$[x] =$ numSupports both before and after the deletion have no implicit support now, but did not lose implicit support because of this deletion.) Fortunately, our existing maintenance of data structures happens to compact exactly these variables into a particular region of varsBySupport, so we can find them very easily and efficiently. The compaction happens through the sequence of calls to the Procedure swap made by Procedure 2. We first show a worked example and we then prove the general properties we need.

In Figure 1, we suppose there are 11 variables in a constraint, there are currently 8 supports, and we are deleting a support involving variables $x_1$, $x_2$, $x_3$, $z_1$, $z_2$ and $z_3$, with the literals deleted in an arbitrary order from top (start) to bottom (finish). Before we start, the $z$ variables already have supportsPerVar = numSupports = 8; variables $x$ and $y$ have supportsPerVar = 7; and variables $w$ have supportsPerVar = 6. As we process literals in deleteSupport, pairs of variables are swapped (marked by boxes in each line) and the boundaries move between cells (marked by vertical lines) of variables with equal supportsPerVar. At the end, $w$ and $x$ variables still have supportsPerVar = 6 < numSupports = 7. The $z$ variables have supportsPerVar=numSupports both before and after deletion. The only variables that have *lost* their *last* implicit support are the $y$ variables. The crucial point is that at the end they lie precisely between the final boundary between 6 and 7 supports (from $i = 5$), and the initial boundary between 7 and 8 supports (from $i = 8$). The following simple results show that variables losing their last implicit support are always compacted in a similar way.

**Lemma 6.1.** *Suppose, before we delete a support $S$, that* numSupports $= p$ *(and so num-Supports $= p - 1$ afterwards). For a variable $x$ to lose its last implicit support, it has $p - 1$ explicit supports both before and after the deletion of $S$.*

*Proof.* If $x$ initially has fewer than $p-1$ explicit supports, then $x$ has more than one implicit support and deleting $S$ removes at most one of these. If $x$ initially has $p$ explicit supports, then it is involved in $S$ (since it is involved in all supports) and so has no implicit support to lose. Hence, $x$ must initially have $p - 1$ explicit supports and one implicit support and $S$ must be that one implicit support. Therefore after the deletion of $S$, $x$ has $p - 1$ explicit supports and no implicit supports. □

**Lemma 6.2.** *We set $p$ as in Lemma 6.1, $i$ as the value of* supportNumLowIdx$[p]$ *when deleteSupport is called, and $j$ as the value of* supportNumLowIdx$[p - 1]$ *when deleteSupport exits. When deleteSupport finishes, the variables that lost their last implicit support during the call to deleteSupport are exactly the set of variables at indices in the range $[j, i)$ in* varsBySupport.

*Proof.* All variables with no implicit supports when deleteSupport exits lie at index $j$ or greater in varsBySupport. This establishes the lower bound on the index range.

Any variable $z$ that has no implicit support at the start of the call must have $p$ explicit supports and so must be at index $i$ or higher. $z$ must be in the support being deleted, because it is in all supports. When $z$ is updated by deleteSupport, it is always swapped with the variable at index supportNumLowIdx$[p]$. The index supportNumLowIdx$[p]$ only increases during deleteSupport, so $z$ stays at index $i$ or higher throughout. Thus the variables from index $i$ upwards at the finish are a permutation of those at the start, meaning that variables which lost their last implicit support must be in the range $[j, i)$. Finally, any variable in the range $[j, i)$ has no implicit support at the end of the call (as it is at index $j$ or above) but had an implicit support at the start (as it is before $i$). Therefore all and only variables which lost their last implicit support lie at indices in the range $[j, i)$. □

From Lemma 6.2, after we run deleteSupport it is trivial to enumerate all variables which have lost their last implicit support as a result. They are exactly the variables varsBySupport$[k]$ for $k = j, j+1, ...i-1$ with $i$ and $j$ as defined in the Lemma. Enumerating this list is the only additional work over that already done by Procedure 2, so we have:

**Corollary 6.3.** *Given a constraint on $n$ variables, the additional work to identify variables which have lost their last implicit support is $O(1)$ for each such variable where there are some, and $O(1)$ if there are none.*

*Proof.* We have already argued the case where there are variables which have lost implicit support. If there are no such variables, there is still $O(1)$ work to check that the range is empty. □

This low level of complexity contrasts very favourably with SHORTGAC. When a support is deleted, Procedure 4 iterates over all variables with **numSupports** explicit supports. In the worst case this is $O(n)$ work even if no variable has lost its last implicit support, compared to the $O(1)$ work that we now have. We now move on to the details of incorporating these optimisations into a full suite of procedures for maintaining GAC.

### 6.3 HaggisGAC: Details

Two issues complicate the implementation of HaggisGAC compared with SHORTGAC. First, the Lemmas above depend on all literals in a support being deleted in a single pass. Therefore, instead of acting immediately on finding a literal with no supports, we keep a list of literals with lost supports for later treatment. Second, we now have two cases in which we might detect lost support – when the lost support is explicit or implicit – compared to the single case in SHORTGAC, where all lost supports are detected in the same way.

We introduce two simple data structures for storing literals and variables that have lost explicit or implicit support as we find them.

**litsLostExplicitSupport** is a set containing literals that have lost their final explicit support and are not supported implicitly.

**varsLostImplicitSupport** is a set containing variables that have lost their final implicit support.

We have to adapt the deleteSupport procedure from Procedure 2. The new version is shown as Procedure 7. When we find a literal which has no explicit support, we immediately check if it has an implicit support instead (Line 8). If it does not, then we add it to the set litsLostExplicitSupport for later processing to find a new support or delete it. Variables which have no implicit support are detected after all literals have been deleted. This is done by lines 15-16, which are justified by Lemma 6.2.

The new propagate procedure is shown in Procedure 8. Like the earlier Procedure 3, we first delete all supports involving the literal to be deleted, but the rest of the procedure is very different. We first iterate through all literals which lost their last explicit support, and then the variables which lost their last implicit support.

For the lost explicit supports, we call HaggisGAC-literalUpdate (Procedure 9). This procedure has no analogue in SHORTGAC, but is straightforward. The only point of interest is that we still check whether a literal is supported, even though it was only added to litsLostExplicitSupport if it was not. The reason is that some support found by an unrelated call to findNewSupport might also support this literal. If so we are done, but if not then Procedure 9 calls findNewSupport. If a new support is found it is added, but if not then we have to prune the literal as being no longer supported.

**Require:** Short Support $sup$
1:  $oldIndex \leftarrow$ supportNumLowIdx[numSupports]
2:  **for all** $(x \mapsto v) \in sup$ **do**
3:      Remove $sup$ from supportListPerLit$[x \mapsto v]$
4:      **if** supportListPerLit$[x \mapsto v] = \{\}$ **then**
5:          detachTrigger$(x,v)$
6:          **if** $(x \mapsto v) \notin$ zeroLits$[x]$ **then**
7:              Add $(x \mapsto v)$ to zeroLits$[x]$
8:          **if** supportsPerVar$[x] =$ numSupports **then**
9:              Add $(x \mapsto v)$ to litsLostExplicitSupport
10:          sPV $\leftarrow$ supportsPerVar$[x]$
11:          swap$(x,$ varsBySupport[sPV])
12:          supportNumLowIdx[sPV] $\leftarrow$ supportNumLowIdx[sPV]+1
13:          supportsPerVar$[x] \leftarrow$ sPV$-1$
14: numSupports--
15: **for all** $i \in \{$supportNumLowIdx[numSupports] $\ldots oldIndex - 1\}$ **do**
16:      Add varsBySupport$[i]$ to varsLostImplicitSupport

Procedure 7: HaggisGAC-DeleteSupport: $(sup)$. One subtlety is that we must add $(x \mapsto v)$ to zeroLits (line 7) even if we also add it to litsLostExplicitSupport (line 9). The only case where this matters is that we seek and find a new implicit support, i.e. not containing $x \mapsto v$, but this is later lost. At the later point Procedure 10 requires $x \mapsto v$ to be in zeroLits because $x \mapsto v$ might still have no explicit support.

**Require:** $x \nmapsto v$ (where $v$ has been pruned from domain of $x$)
1: litsLostExplicitSupport $\leftarrow \{\}$
2: varsLostImplicitSupport $\leftarrow \{\}$
3: **while** supportListPerLit$[x \mapsto v] \neq \{\}$ **do**
4:      $sup \leftarrow$ first element of supportListPerLit$[x \mapsto v]$
5:      deleteSupport$(sup)$
6: **for all** $(y \mapsto b) \in$ litsLostExplicitSupport **do**
7:      HaggisGAC-literalUpdate$(y \mapsto b)$
8: **for all** $z \in$ varsLostImplicitSupport **do**
9:      HaggisGAC-variableUpdate$(z)$

Procedure 8: HaggisGAC-Propagate: propagate$(x \nmapsto v)$

For variables with lost implicit supports, we call HaggisGAC-variableUpdate (Procedure 10), which is similar to Procedure 4. The differences are that the return statements from Procedure 4 are omitted; we check at every iteration whether a new implicit support has been found for $x$ and if so exit the loop; and we do not remove $x \mapsto v$ from zeroLits if a new explicit support has been found, allowing this to be done lazily in a later call at Line 5.

We gain efficiency over ShortGAC for two reasons. First, variableUpdate is only called for variables that have just lost implicit support. Second, there is no outer loop in HaggisGAC-Propagate which must be restarted when a new support is found, as there is in Procedure 3. If we write $m$ for the number of variables which have lost their last implicit support, we have reduced the worst case number of calls to variableUpdate from HaggisGAC-Propagate from $O(n^2d)$ where $n$ is the arity of the constraint to $m$. Since $m \leq n$ and $m$ can often be much smaller than $n$ or even zero, this is a significant gain.

**Require:** $x \mapsto v$, where last explicit support of $x \mapsto v$ has been deleted
1: **if** $v \in D(x)$ **and** supportsPerVar$[x]$ = numSupports **and**
    supportListPerLit$[x \mapsto v] = \{\}$ **then**
2:    $sup \leftarrow$ findNewSupport$(x, v)$
3:    **if** $sup = $ Null **then**
4:       prune$(x \mapsto v)$
5:    **else**
6:       addSupport$(sup)$

Procedure 9: HaggisGAC-literalUpdate$(x \mapsto v)$

**Require:** variable $x$
1: **for all** $(x \mapsto v) \in$ zeroLits$[x]$ **do**
2:    **if** supportsPerVar$[x] <$ numSupports **then**
3:       **return**
4:    **if** supportListPerLit$[x \mapsto v] \neq \{\}$ **then**
5:       Remove $(x \mapsto v)$ from zeroLits$[x]$
6:    **else**
7:       **if** $v \in D(x)$ **then**
8:          $sup \leftarrow$ findNewSupport$(x \mapsto v)$
9:          **if** sup = Null **then**
10:            prune$(x \mapsto v)$
11:          **else**
12:            addSupport(sup)

Procedure 10: HaggisGAC-variableUpdate$(x)$

## 6.4 Dealing Efficiently With Full-length Supports

When a full-length support is added, ShortGAC increments numSupports and supports-PerVar for every variable. Since we are only interested in the condition numSupports = supportsPerVar$[x]$, a full-length support cannot change this status for any variable. Therefore we can save overheads in the case where we add a full-length support. This is achieved through a case split in HaggisGAC's versions of addSupport and deleteSupport: if a support is full length we do not update numSupports, supportsPerVar, and related data structures. Note that the test we apply is not that the final support is of arity $n$, but the initial one before the omission of any assigned literals as the optimisation is correct even if assigned literals are omitted. We omit the pseudocode for this optimisation, as the changes are straightforward. This optimisation often improves performance on instances with all full-length supports by 20%, and has no important effect on our other instances with runtimes all within $\pm 2.5\%$ with or without it. This optimisation is also applicable to ShortGAC, but we did not implement it in that case because it does not address the key inefficiency that algorithm has, i.e. the repeated checking of variables which cannot have lost their last implicit support. This does not affect our experimental results dramatically: in most cases we found that the improved performance of HaggisGAC was larger than this optimisation provides.

## 7. Experimental Evaluation of SHORTGAC and HAGGISGAC

The Minion solver 0.12 (Gent, Jefferson, & Miguel, 2006a) was used for our experiments, with the only changes being the additional propagators. In all experiments, all the compared methods maintain GAC. Therefore, the solver explores the same search space in each case. Since the number of nodes searched is invariant, we compare the rate of search exploration, measured in search nodes per second.[6]

We used an 8-core machine with 2.27GHz Intel Xeon E5520 CPUs and 12GB memory, running Ubuntu Linux. Where possible we ran 12 processes in parallel. For each combination of problem instance and propagator, we report the median of 11 runs.[7] In some cases it is not possible to run 12 processes in parallel because they exceed 1GB memory. For these, we ran just one process at a time, and we report the median of 5 runs. These instances are marked with a '‡' in the tables of results. If one method exceeded 1GB, we sometimes ran other comparable methods in series as well. This allows consistent comparison between List and NDList, and different propagation algorithms. It also means that '‡' in the tables does not necessarily indicate that the method uses more than 1GB memory. We find the median to be a very robust measure of performance, for reasons described in Appendix B.

In all cases, we imposed a time limit of one hour, and a limit of 1,000,000 search nodes (whichever is first). To avoid short runs when the solver can find a solution easily, we searched for all solutions. We report complete cpu times, i.e. we have not attempted to measure the time attributable to the given propagator and we include any initialisation. This has the advantage that we automatically take account of all factors affecting runtime, including aspects (e.g. cache usage) that we may not realise affect runtime. It does however mean that our results tend to *understate* the difference between methods being studied.

For each case study, we implemented a findNewSupport method for SHORTGAC and HAGGISGAC specific to the constraint. We also used the generic list instantiation (Section 5.5.1) and the Next-Difference List instantiation (Section 5.5.2) for comparison where possible. We compare SHORTGAC and HAGGISGAC with the special-purpose propagator (when available).

We also compare with SHORTGAC-Long (SHORTGAC with full-length supports), with HAGGISGAC-Long, and with GAC-Schema (Bessière & Régin, 1997) as the closest equivalent algorithm without strict short supports. We discuss GAC-Schema further in Section 7.4. GAC-Schema, SHORTGAC-Long and HAGGISGAC-Long use the same (constraint-specific) findNewSupport as SHORTGAC, and subsequently extend the short support to full length using the minimum value for each extra variable.

In each case, the constraint can be compactly represented as a disjunction. Therefore we compare SHORTGAC and HAGGISGAC with Constructive Or. The algorithm used is based on Lagerkvist and Schulte's (2009), without the rule for entailment detection. The implementation in Minion is fully incremental: each disjunct is propagated incrementally down a branch of search and backtracked as the search backtracks.[8]

---

6. Source code for the solver with the three algorithms is available at `http://www.cs.st-andrews.ac.uk/~pn/haggisgac-source.tgz` and problem instances and experimental results at `http://www.cs.st-andrews.ac.uk/~pn/haggisgac-data-instances.tgz`.
7. In preliminary investigations, we found that running 12 processes in parallel gives consistent cpu time results, and this consistency is improved by taking the median.

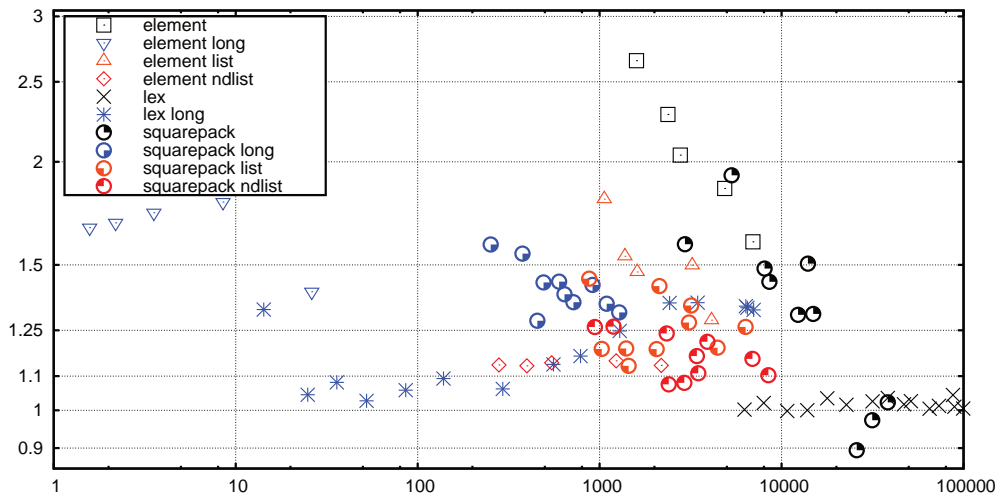

Figure 2: Summary comparison of ShortGAC and HaggisGAC. The $x$-axis is median nodes per second for ShortGAC. The $y$-axis is speedup (or slowdown) of HaggisGAC, i.e. the ratio of ShortGAC nodes per second to that of HaggisGAC. Hence 1 represents equal behaviour, while above 1 means that HaggisGAC was faster.

We do not compare with table constraints, as described by (for example) Gent et al. (2007), because the constraints are too large. For example, the smallest element constraints reported below have $6^{38}$ allowed tuples, making it impossible even to generate and store the list of allowed tuples.

To aid comparison between HaggisGAC and ShortGAC, in addition to the tables we compare them graphically in Figure 2. This figure shows the relative speedup (or in some cases slowdown) of using HaggisGAC compared with ShortGAC.

## 7.1 Case Study 1: Element

We use the quasigroup existence problem QG3 (Colton & Miguel, 2001) to evaluate Short-GAC and HaggisGAC on the element constraint. The problem class has one parameter $n$, specifying the size of an $n \times n$ table ($qg$) of variables with domains $\{0 \ldots n-1\}$. Rows, columns and one diagonal have GAC allDifferent constraints, following Colton and Miguel's model. The element constraints represent the QG3 property that $(i * j) * (j * i) = i$ (where $i$ and $j$ are members of the quasigroup and $*$ is the quasigroup operator). This translates as $\forall i, j :$ element($qg$, aux$[i, j]$, $i$), and aux$[i, j]$= $n \times qg[i, j] + qg[j, i]$, where aux$[i, j]$ has domain $\{0 \ldots n \times n - 1\}$.

---

8. Personal communication with Pascal Van Hentenryck indicated that there is an unpublished optimisation of Constructive Or whereby some disjuncts need not be propagated in some cases. We did not implement this optimisation.

| $n$ | Watch Elt. | ShortGAC | | | | HaggisGAC | | | | GAC Sch. | Con Or |
|---|---|---|---|---|---|---|---|---|---|---|---|
| | | Specific | List | NDL | Long | Specific | List | NDL | Long | | |
| 6 | 27,825 | 6,956 | 4,122 | 2,182 | 25.9 | 11,131 | 5,300 | 2,473 | 36.5 | 22.5 | 53.5 |
| 7 | 22,259 | 4,866 | 3,226 | 1,233 | ‡8.5 | 9,035 | 4,833 | 1,415 | ‡15.2 | 7.1 | 24.2 |
| 8 | 15,635 | 2,773 | 1,609 | 545 | ‡3.6 | 5,652 | 2,367 | 622 | ‡6.2 | ‡3.0 | 9.1 |
| 9 | 15,898 | 2,374 | 1,377 | 398 | ‡2.2 | 5,419 | 2,116 | 451 | ‡3.7 | ‡3.0 | ‡6.2 |
| 10 | 15,088 | ‡1,594 | ‡1,060 | ‡280 | ‡1.6 | ‡4,227 | ‡1,911 | ‡317 | ‡2.6 | mem | ‡4.2 |

Table 1: Nodes searched per second for quasigroup existence problems. 'mem' indicates running out of memory (>12 GB). Columns correspond to propagation algorithms. Watch Elt is the special-purpose propagator. Both ShortGAC and HaggisGAC have four instantiations: Specific (special-purpose findNewSupport function for the constraint), List, NDL (Next-Difference List), and Long (as described in the text). GAC-Sch is GAC-Schema, and Con Or is Constructive Or.

For the constraint element$(X, y, z)$, the findNewSupport method for ShortGAC returns tuples of the form $\langle x_i \mapsto j, y \mapsto i, z \mapsto j \rangle$, where $i$ is an index into the vector $X$ and $j$ is a common value of $z$ and $x_i$. ShortGAC-list has all supports of this form. For Constructive Or, we used $(x_0 = z \wedge y = 0) \vee (x_1 = z \wedge y = 1) \vee \cdots$.

We compare ShortGAC and HaggisGAC with the special-purpose Watched Element propagator (Gent et al., 2006b), GAC-Schema and Constructive Or. Table 1 presents our results on QG3. Of the general purpose methods, using short supports (with Specific, List or NDList instantiations) is dramatically better than any alternative. For example at $n = 10$, even the HaggisGAC-List method (which is slower than HaggisGAC-Specific) is more than 450 times faster than Constructive Or, the best of the other methods.

ShortGAC-Long runs about 10–20% faster than GAC-Schema for $n = 6$ to 8, slower at $n = 9$ but better at $n = 10$ because GAC-Schema uses more memory. Recall that they both use the same findNewSupport method, so this is a fair comparison of how efficiently they exploit these supports. This is in contrast to our results reported previously (Nightingale et al., 2011), where ShortGAC was about half the speed of GAC-Schema. Two substantial differences account for the improvement: the improved data structures described in Section 5; and that we remove assigned literals from the full-length supports as described in Section 5.6. HaggisGAC-Long is consistently faster than both ShortGAC-Long and GAC-Schema.

While much faster than methods using full-length supports, list variants HaggisGAC-List and HaggisGAC-NDList are both slower than HaggisGAC-Element (and the same is true for ShortGAC). This is to be expected as neither is specialised to the Element constraint, and both have to deal with data structures containing the lists of tuples. Of the two list variants, the NDList variant runs much more slowly. However, its memory usage is, as we expected, much less than HaggisGAC-List. It used less than half as much memory at $n = 6$, improving to almost 10 times less memory at $n = 10$.

HaggisGAC-Element is approximately twice as fast as ShortGAC-Element on these instances. We believe this is because two variables are in all short supports – the index and result variables – meaning that they are always supported explicitly. As can be seen in Figure 2, List, NDList and Long instantiations of HaggisGAC are also faster than the same instantiations of ShortGAC but by a smaller margin. The special purpose Watched Element propagator is the fastest method, being 3.6 times faster when $n = 10$. Watched Element also appears to be scaling better as $n$ increases. Constructive Or is much slower than all the methods that exploit strict short supports, however it is faster than HaggisGAC-Long. Overall it is clear that exploiting strict short supports is very beneficial compared with other general purpose methods.

| $n$ | GACLex | ShortGAC | | HaggisGAC | | GAC-Schema | Con Or |
|---|---|---|---|---|---|---|---|
| | | Specific | Long | Specific | Long | | |
| 3 | 104,955 | 87,463 | 7,020 | 91,265 | 9,288 | 3,622 | 5,735 |
| 4 | 103,950 | 99,602 | 6,481 | 100,100 | 8,628 | 3,030 | 4,997 |
| 5 | 95,420 | 89,127 | 6,358 | 90,009 | 8,503 | 2,734 | 4,104 |
| 6 | 80,841 | 73,260 | 3,456 | 74,184 | 4,666 | 1,638 | 2,109 |
| 7 | 72,307 | 65,062 | 2,424 | 65,359 | 3,271 | 1,190 | 1,188 |
| 8 | 66,445 | 51,335 | 1,290 | 52,659 | 1,609 | 670 | 456 |
| 9 | 64,267 | 47,059 | 786 | 47,847 | 914 | 451 | 263 |
| 10 | 57,208 | 38,344 | 557 | 39,683 | 634 | 318 | 184 |
| 12 | 48,146 | 31,626 | 293 | 32,425 | 311 | 170 | 105 |
| 14 | 36,751 | 22,712 | 139 | 23,063 | 142 | 82.3 | [‡]99.1 |
| 16 | 30,057 | 17,813 | 85.9 | 18,420 | 90.9 | 51.5 | [‡]62.6 |
| 18 | 22,432 | 13,843 | 52.4 | 13,845 | 53.8 | 33.3 | [‡]48.3 |
| 20 | 16,625 | 10,734 | 35.9 | 10,711 | 38.9 | 21.0 | [‡]36.7 |
| 22 | 12,450 | 7,976 | 24.9 | 8,141 | 26.0 | 12.5 | [‡]27.0 |
| 24 | 9,526 | 6,255 | 14.3 | 6,268 | 18.9 | [‡]7.3 | [‡]21.8 |

Table 2: Nodes searched per second for BIBDs. GACLex is the special-purpose propagator, and other columns are named as in Table 1.

## 7.2 Case Study 2: Lex-ordering

We use the BIBD problem to evaluate ShortGAC and HaggisGAC on the lexicographic ordering constraint. The lex constraint is placed on both the rows and columns, to perform the 'Double Lex' symmetry breaking method (Flener et al., 2002). We use the BIBD model given by Frisch, Hnich, Kiziltan, Miguel, and Walsh (2002), and the GACLex propagator given by Frisch, Hnich, Kiziltan, Miguel, and Walsh (2006). We use BIBDs with the parameter values $(4n + 3, 4n + 3, 2n + 1, 2n + 1, n)$.

For the constraint lexleq$(X, Y)$ on arrays $X$ and $Y$, we define $mx_i = min(Dom(x_i))$ and $my_i = max(Dom(y_i))$. The findNewSupport method for ShortGAC finds the lowest index $i \in \{0 \dots n\}$ such that $mx_i < my_i$, or $i = n$. The case $i = n$ arises when $X$ cannot be lexicographically less than $Y$, so a support is sought for $X = Y$. If $i < n$, the support contains $x_i \mapsto mx_i$, $y_i \mapsto my_i$. For each index $j < i$, if $mx_j = my_j$, then the short support contains $x_j \mapsto mx_j, y_j \mapsto my_j$ otherwise there is no valid support and Null is returned.

The lex constraint on two arrays of length $n$ and domain size $d$ has more than $d^n$ short supports in any short support set, because all assignments where the two arrays are equal satisfy the constraint and cannot be reduced. ShortGAC-List and ShortGAC-NDList are not practical for any substantial constraint so we omit them from the comparison. For Constructive Or we use the following representation with $n + 1$ disjuncts: $(x_0 < y_0) \lor (x_0 = y_0 \land x_1 < y_1) \lor \cdots$, including the final case where all pairs are equal.

Table 2 presents the results of our experiments on non-list based methods with values of $n$ from 3 to 24. It is clear that the best method is the special-purpose GACLex propagator, with HaggisGAC coming second. On this problem, HaggisGAC and ShortGAC perform similarly. HaggisGAC and ShortGAC are by far the best general purpose methods. For the largest instances they run about 1.5 times slower than the special purpose method, while outperforming the next best method by almost 300 times. Again, HaggisGAC-Long and ShortGAC-Long outperform GAC-Schema, and on these instances the difference is even more marked.

HaggisGAC-Long can be substantially faster than ShortGAC-Long, as can be seen in Figure 2: this is largely explained by the optimisation of Section 6.4.

To summarise, these experiments on the Lex constraint clearly show the benefit of HaggisGAC and ShortGAC compared with other general-purpose propagation methods. Their speed even approaches that of the special purpose GACLex propagator.

## 7.3 Case Study 3: Rectangle Packing

The rectangle packing problem (Simonis & O'Sullivan, 2008) (with parameters $n$, *width* and *height*) consists of packing all squares from size $1 \times 1$ to $n \times n$ into the rectangle of size *width* $\times$ *height*. This is modelled as follows: we have variables $x_1 \ldots x_n$ and $y_1 \ldots y_n$, where $(x_i, y_i)$ represents the Cartesian coordinates of the lower-left corner of the $i \times i$ square. Domains of $x_i$ variables are $\{0 \ldots width - i\}$, and for $y_i$ variables are $\{0 \ldots height - i\}$. Variables are branched on in decreasing order of $i$ (to place the largest square first), with $x_i$ before $y_i$, smallest value first. The only type of constraint is non-overlap of squares $i$ and $j$: $(x_i + i \leq x_j) \lor (x_j + j \leq x_i) \lor (y_i + i \leq y_j) \lor (y_j + j \leq y_i)$. Minion does not have the special-purpose non-overlap constraint (Simonis & O'Sullivan, 2008), so we only report a comparison of general-purpose methods. For the experiment we used the optimum rectangle sizes reported by Simonis and O'Sullivan.

The domains of $x_n$ and $y_n$ are reduced to break flip symmetries as described by Simonis and O'Sullivan (2008). Our focus is performance of the non-overlap constraint, and so we did not implement the commonly-used implied constraints.

The findNewSupport function for ShortGAC is as follows. If any of the four disjuncts above are entailed given the current domains, return the empty support (indicating entailment). Otherwise, return a support with two literals to satisfy one of the four disjuncts. The list used for ShortGAC-List and ShortGAC-NDList has all supports of size 2.

In Table 3, we compare HaggisGAC and ShortGAC with other general purpose methods. We can see that HaggisGAC is the fastest method, with ShortGAC second. HaggisGAC-List and HaggisGAC-NDList (as well as ShortGAC-List and ShortGAC-NDList) performed well compared to GAC-Schema and Constructive Or. However at $n = 20$, HaggisGAC-List consumes 971MB memory and HaggisGAC-NDList 496MB, and with $n > 20$ it was not possible to run these methods with 12 processes in parallel. Interestingly, the performance of the two List variants of HaggisGAC is reversed from Case Study 1: here, NDList is significantly faster than List in most cases. As expected, NDList used less memory, though less dramatically than before. It used from about 30% to 50% of the memory of HaggisGAC-List.

| | ShortGAC | | | | HaggisGAC | | | | GAC | Con |
|---|---|---|---|---|---|---|---|---|---|---|
| *n-w-h* | Specific | List | NDL | Long | Specific | List | NDL | Long | Sch. | Or |
| 18-31-69 | 14,923 | 6,339 | 6,919 | 1,093 | 19,524 | 7,999 | 7,988 | 1,471 | 1,033 | 441 |
| 19-47-53 | 38,329 | 4,446 | 8,460 | 1,282 | 39,185 | 5,295 | 9,330 | 1,684 | 1,181 | 478 |
| 20-34-85 | 13,949 | 3,181 | 3,911 | 914 | 21,000 | 4,261 | 4,734 | 1,296 | 775 | 276 |
| 21-38-88 | 8,568 | ‡2,668 | ‡2,781 | 641 | 12,262 | ‡3,955 | ‡3,981 | 886 | 592 | 245 |
| 22-39-98 | 8,059 | ‡1,865 | ‡1,889 | 599 | 11,966 | ‡3,013 | ‡2,896 | 858 | 518 | 185 |
| 23-64-68 | 31,486 | ‡1,226 | ‡2,805 | 718 | 30,628 | ‡1,663 | ‡3,863 | 971 | 590 | 349 |
| 24-56-88 | 12,317 | ‡1,717 | ‡2,238 | 492 | 16,075 | ‡2,441 | ‡3,152 | 702 | 474 | 167 |
| 25-43-129 | 5,310 | ‡1,007 | ‡986 | 377 | 10,228 | ‡1,634 | ‡1,506 | 583 | 348 | 96 |
| 26-70-89 | 25,860 | ‡909 | ‡1,977 | 455 | 23,132 | ‡1,219 | ‡2,577 | 584 | 376 | 245 |
| 27-47-148 | 2,943 | ‡1,034 | ‡786 | 252 | 4,677 | ‡1,265 | ‡1,187 | 400 | 272 | 74 |

Table 3: Nodes searched per second for Rectangle Packing instances. All columns are named as in Table 1.

Of the other methods, all are always at least 10 times slower than HaggisGAC. HaggisGAC-Long is faster than GAC-Schema in all cases. Also ShortGAC-Long is faster than GAC-Schema on all instances except 27-47-148 (this contradicts the result we previously reported (Nightingale et al., 2011), and some explanation of this is given in the first case study).

Table 3 shows that HaggisGAC (with the SquarePack instantiation) is substantially faster than ShortGAC on most of the instances, with the exception of $n = 23$ and $n = 26$ where ShortGAC is slightly faster. When compared with ShortGAC for List, NDList, and Long instantiations in Figure 2, we see that HaggisGAC is mostly between 10 and 50% faster. In summary, these results very clearly show the benefits of using strict short supports.

### 7.4 Comparing HaggisGAC With GAC-Schema

Across all the above experiments, HaggisGAC-Long runs significantly faster than GAC-Schema – from a minimum of about 20% faster to more than three times faster – even though our code contains overhead for dealing with strict short supports. We compared memory usage across all experiments, and found very similar performance across all instances. We found that HaggisGAC-Long uses less than 5% more memory on all except BIBD instances, and on BIBD it uses less than 17% more memory than GAC-Schema.

However, the comparison has been only on functional instantiations of full-length supports, and on constraints that admit strict short supports. In this section, we broaden the comparison by using the list instantiations rather than functional ones, and using problem instances that have been used previously for comparing *table* constraints.

We compared against GAC-Schema because it is very similar to HaggisGAC and ShortGAC conceptually. All three algorithms maintain a list of supports for each literal, which is updated and backtracked during search. GAC-Schema was carefully implemented following the pseudocode of the original paper (Bessière & Régin, 1997). While some code is shared among all three algorithms, each was optimised independently. For example, GAC-Schema has a different implementation of supportListPerLit, named $S_C$ in (Bessière & Régin, 1997), which is specialised to full-length supports.

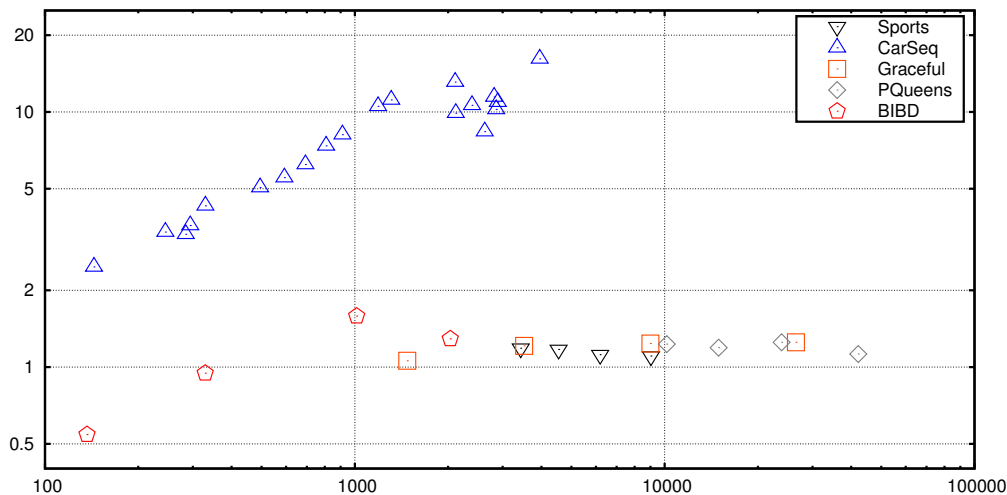

Figure 3: Comparison of GAC-Schema and HaggisGAC-List on full-length table constraints. $x$-axis is nodes per second for GAC-Schema, $y$-axis is speedup of HaggisGAC-List.

In contrast to GAC-Schema, other table constraint propagators such as STR2 (Lecoutre, 2011) and MDDC (Cheng & Yap, 2010) are entirely different to HaggisGAC, and it would be difficult to create truly comparable implementations of them.

We report on the use of HaggisGAC-List only, because it searches for supports in the same way as GAC-Schema (with one difference we discuss below.) We used the structured instances from Gent et al. (2007), except the Semigroup class. In addition, we used Car Sequencing instances from Nightingale (2011), specifically model B instances numbered 60-79. These instances contain a large number of ternary table constraints.

Figure 3 shows that HaggisGAC-List is almost always faster than GAC-Schema on these problems. For BIBDs it is not clear which algorithm is better. HaggisGAC is always at least marginally faster on the Sports Scheduling, Prime Queens and Graceful Graphs instances, in most cases in the range 10-20% faster. HaggisGAC is substantially faster on Car Sequencing. To seek new supports, HaggisGAC calls Procedure 5, and when it finds a new support it stores the index of it in listPos. HaggisGAC does not backtrack listPos as described in Section 5.5.1. GAC-Schema is similar, but it does backtrack listPos, and it ensures optimality down a branch of search by iterating only from listPos to the end of the list (Bessière & Régin, 1997). Profiling shows that GAC-Schema is hindered by backtracking listPos (by block-copying memory) on Car Sequencing, where there are a very large number of table constraints (2000 on instance 60) and large domains (some of size over 1000). Alternative memory management techniques might speed up GAC-Schema, so we do not claim that HaggisGAC is fundamentally 10 times faster than GAC-Schema.

## 7.5 Results Summary

To summarise the three case studies, HaggisGAC does indeed outperform ShortGAC on many instances, sometimes by more than two times and commonly by more than 25%. ShortGAC is only rarely faster, but only on one instance by as much as 10%. Overall, in our experiments, HaggisGAC is clearly a better algorithm than ShortGAC. Furthermore, HaggisGAC and ShortGAC perform very well compared to Constructive Or and GAC-Schema, a result which validates the idea of strict short supports.

Finally, we have shown experimentally that HaggisGAC can outperform GAC-Schema on problems containing only full-length supports. We discuss why this should be in Appendix C as it is not a major focus of this paper.

## 8. Backtrack Stability and Short Supports

Within a search tree, HaggisGAC often spends significant time backtracking data structures. Reducing or eliminating backtracking can improve efficiency. For example MAC-6 and MAC-7 can be much more efficient (in both space and time) if backtracking is avoided (Régin, 2005). In this section we present a new algorithm that saves time by not deleting short supports on backtrack, and saves memory by bounding the total number of stored short supports (including those on the backtrack stack).

The new algorithm requires that short supports have the backtrack stability property. A short support is backtrack stable iff it remains a short support after backtracking (Section 3.2).

In our three case studies, we find that the short supports we construct for the element and lex constraints are backtrack stable, but for rectangle packing they are not. For rectangle packing, we generate the empty support when the constraint is entailed. The empty support is not backtrack stable unless the constraint is entailed at the root node of search.

We introduce the algorithm HaggisGAC-Stable where we know all short supports are backtrack stable. The key change is that we do not delete supports when we backtrack past their point of introduction. Because they are stable, they are still correct at ancestors of the node they were introduced at. This can save time over the previous algorithms, since we sometimes need to do no work at all on backtracking. Also, as we show below, we obtain very tight limits on space usage of stored supports.

To present HaggisGAC-Stable, we introduce the notion of a *prime support* of a deleted literal. A prime support of a deleted literal is a support (either explicit or implicit) which will be a valid support for that literal when the literal is restored on backtracking. The invariant we maintain after deleting a literal is that *either* we have labelled a deleted support on the backtrack stack as its prime support *or* the literal's variable is currently implicitly supported. With this invariant, we guarantee that when we backtrack to the point where the literal is restored, it must be supported again: either by the prime support which we can restore, or by the known implicit support.

The task of finding the prime support for a literal naturally splits into three cases. The simplest case is that HaggisGAC-Stable itself deletes literals when not able to find a necessary new support. The prime support is then just the implicit or explicit support whose deletion caused the fruitless search for a new support.

The second case is where a literal is pruned by some other constraint or the search procedure, and the pruned literal had an explicit support in this constraint. All its explicit supports must be deleted as no longer valid, and we can label an arbitrary one to be the literal's prime support: we simply choose the last one to be deleted.

The third case is unfortunately complicated. It is that a literal is pruned outside the current constraint, and the literal had an implicit support but no explicit support. This is difficult precisely because the pruned literal does not have any link to its implicit support. Providing and maintaining such a link throughout search would negate the efficiencies we have gained. Our solution to this problem is to be lazy. The variable of the pruned literal is implicitly supported. While we have any implicit support for the variable, we are maintaining the invariant described above. So when the literal is pruned we need do nothing in this case. We only need do any work when this variable loses its last implicit support, if it ever does. When this happens, an invalid literal which had no explicit support must by definition be in the relevant **zeroLits** list. Whereas previously we ignored invalid literals when iterating through **zeroLits**, we now can label the deleted implicit support as a prime support for the invalid literal.

We will show in Lemma 8.1 that HaggisGAC-Stable stores at any time at most $O(z)$ supports, where $z$ is the total number of literals. This can save a lot of memory because HaggisGAC and ShortGAC may store $O(z^2)$ supports, because there can be $O(z)$ deletions of literals down a branch, and for each deletion a new set of $O(z)$ supports may be stored. Our experiments later will show that this difference in memory usage can be significant in practice. At its most effective, memory usage was reduced by 20 times.

## 8.1 Details of HaggisGAC-Stable

In HaggisGAC-Stable, we have to control with great care the deletion and restoration of supports, instead of (as in the rest of this paper) simply reversing the addition or deletion of a support at a node by respectively deleting or adding it back when we backtrack past that node. In short we *never* delete an active support on backtracking, and *only* add back in a deleted support if it is a prime support for a literal with no current active support.

When deleting a support, we setup a counter **numPrimeSupported**. It is initially 0, and is incremented each time we find the support is a prime support. When the propagation algorithm finishes, for any support with **numPrimeSupported** = 0, the support can be destroyed and its space reclaimed. Otherwise, we place **numPrimeSupported** new pairs on the backtrack stack. Each pair consists of the deleted support and the literal it is a prime support for. On backtracking, when we pop a pair, we first check if any current support already supports the literal. If so, we simply decrement **numPrimeSupported**, and if this reduces to 0, again we reclaim the support's space. If the literal is not supported, then we restore the support via a call to addSupport. In this way all literals the support was prime for are now guaranteed to be supported.

A relatively minor difference is that when we iterate through **zeroLits** we now delete invalid literals from **zeroLits**. We can do this because on backtracking we can restore them into

**Require:** $x \mapsto v$, where last explicit support of $x \mapsto v$ has been deleted
 1: **if** $v \in D(x)$ **then**
 2:    **if** supportsPerVar$[x]$ = numSupports **and** supportListPerLit$[x \mapsto v] = \{\}$ **then**
 3:       sup ← findNewSupport$(x, v)$
 4:       **if** sup = Null **then**
 5:          prune$(x \mapsto v)$
 6:          **increment** lastSupportPerLit$[x \mapsto v]$.numPrimeSupported
 7:          **push** $\langle x \mapsto v,$ lastSupportPerLit$[x \mapsto v] \rangle$ onto BacktrackStack
 8:       **else**
 9:          addSupport(sup)
10: **else**
11:    **increment** lastSupportPerLit$[x \mapsto v]$.numPrimeSupported
12:    **push** $\langle x \mapsto v,$ lastSupportPerLit$[x \mapsto v] \rangle$ onto BacktrackStack zeroLits because they are on the backtrack stack, and doing so enables the space complexity result in Lemma 8.1.

Procedure 11: HaggisGAC-Stable-literalUpdate: $(x \mapsto v)$. In comparison to Procedure 9, we update numPrimeSupported and BacktrackStack.

HaggisGAC-Stable is similar to HaggisGAC. Where appropriate we simply describe differences to save space. The Procedure HaggisGAC-Stable-Propagate is almost the same as Procedure 8, calling the backtrack stable variants of deleteSupport, literalUpdate (Procedure 11) and variableUpdate (Procedure 12). In addition, at the end of this algorithm we destroy and reclaim the space for any deleted support for which numPrimeSupported = 0. The Procedure HaggisGAC-Stable-DeleteSupport (called with support $S$) is also very similar to its predecessor, Procedure 7, with some additions. First, it initialises numPrimeSupported for $S$ to 0. Second, we have new data structures lastSupportPerLit for a deleted literal $x \mapsto a$ and lastSupportPerVar for a variable $x$. In terms of Procedure 7, these are both assigned to $S$ at Line 9 and Line 16 (respectively). Note these assignments do *not* make $S$ a prime support: this will be checked later.

Procedure 11 is analogous to Procedure 9 but with enough differences that we show it in detail here. It identifies prime supports, and when necessary increments numPrimeSupported and pushes invalid literal/support pairs onto the backtrack stack. We also present Procedure 12 in detail, the analogue to Procedure 10. Again it identifies prime supports, increments the counter and adds pairs to BacktrackStack. One difficult case arises, from Line 17. Here, $x \mapsto a$ has been pruned, but externally to this constraint. If it had been pruned by Procedure 11, it would not be in zeroLits. When $x \mapsto a$ is restored on backtracking we still need to make sure it has support. Since it has no explicit support (it is in zeroLits), its last support must be this implicit support we are deleting. Therefore we store the support on BacktrackStack. A minor change to note is that we remove literals from zeroLits, at Lines 13 and 19.

Whenever a new search node (including the root) is entered, a Null is pushed onto the BacktrackStack. This is used as a marker for the procedure HaggisGAC-Stable-Backtrack (Procedure 13), which processes literal/support pairs until it reaches the Null. This restores prime supports for literals being put back into the domain on backtracking, but only if no other support is currently known. If the numPrimeSupported counter for

**Require:** variable $x$
1: **for all** $(x \mapsto v) \in \mathsf{zeroLits}[x]$ **do**
2:  **if** $\mathsf{supportsPerVar}[x] < \mathsf{numSupports}$ **then**
3:   **return**
4:  **if** $\mathsf{supportListPerLit}[x \mapsto v] \neq \{\}$ **then**
5:   Remove $(x \mapsto v)$ from $\mathsf{zeroLits}[x]$
6:  **else**
7:   **if** $v \in D(x)$ **then**
8:    $sup \leftarrow \text{findNewSupport}(x, v)$
9:    **if** $sup = \text{Null}$ **then**
10:     $\text{prune}(x \mapsto v)$
11:     **increment** $\mathsf{lastSupportPerVar}[x].\mathsf{numPrimeSupported}$
12:     **push** $\langle x \mapsto v, \mathsf{lastSupportPerVar}[x] \rangle$ onto $\mathsf{BacktrackStack}$
13:     Remove $(x \mapsto v)$ from $\mathsf{zeroLits}[x]$
14:    **else**
15:     $\text{addSupport}(sup)$
16:   **else**
17:    **increment** $\mathsf{lastSupportPerVar}[x].\mathsf{numPrimeSupported}$
18:    **push** $\langle x \mapsto v, \mathsf{lastSupportPerVar}[x] \rangle$ onto $\mathsf{BacktrackStack}$
19:    Remove $(x \mapsto v)$ from $\mathsf{zeroLits}[x]$

Procedure 12: HaggisGAC-Stable-variableUpdate: $(x)$. This is similar to Procedure 10 with the addition of maintenance of $\mathsf{numPrimeSupported}$ and $\mathsf{BacktrackStack}$.

1: **while** the top element of $\mathsf{BacktrackStack}$ is not Null **do**
2:  **pop** $\langle x \mapsto v, sup \rangle$ from $\mathsf{BacktrackStack}$
3:  **if** $sup$ has not yet been restored **then**
4:   **if** $\mathsf{supportsPerVar}(x) = \mathsf{numSupports}$ **and** $\mathsf{supportListPerLit}[x \mapsto v] = \{\}$ **then**
5:    HaggisGAC-Stable-AddSupport$(sup)$
6:   **else**
7:    {Another support exists for $x \mapsto v$}
8:    **decrement** $sup.\mathsf{numPrimeSupported}$
9:    **if** $sup.\mathsf{numPrimeSupported} = 0$ **then**
10:     **destroy** $sup$ and reclaim space
11:  **if** $\mathsf{supportListPerLit}[x \mapsto v] = \{\}$ **then**
12:   Add $(x \mapsto v)$ to $\mathsf{zeroLits}[x]$
13: **pop** Null from $\mathsf{BacktrackStack}$

Procedure 13: HaggisGAC-Stable-Backtrack. Performs backtracking using $\mathsf{Backtrack}$-$\mathsf{Stack}$.

a support becomes zero, the support can be destroyed as it is no longer necessary. Note that literals are put back into $\mathsf{zeroLits}$ if necessary at Line 12, reversing their deletion in Procedure 12.

We cannot use the optimisation described in Section 5.6, of deleting literals in supports for variables that are assigned, because this may break the backtrack stability property.

However, we retain the optimisation of Section 6.4 for full-length supports, but again omit pseudocode showing this in the interest of focusing on the essential aspects of the algorithms.

## 8.2 Improved Space Complexity of HaggisGAC-Stable

Our approach improves the space complexity of HaggisGAC-Stable compared with HaggisGAC, as the following lemma shows.

**Lemma 8.1.** *For a constraint involving $z$ literals, at most $2z$ supports are stored, either as active or as deleted supports on the backtrack stack.*

*Proof.* We define a function from supports to literals. If the support is still active, it was found from a call to findNewSupport for a specific literal, and we map the support to this literal. Similarly, if the support is on the backtrack stack, then it is in a pair with at least one literal it is a prime support for. Map the support to any one of these literals. Every stored support falls into one of these two categories, because if a support is deleted and it is not put onto the backtrack stack, its space is reclaimed. No three supports are mapped to the same literal because:

- For valid literals, findNewSupport will not be called again if an existing active support exists for that literal.

- For invalid literals, each literal appears in a pair on the backtrack stack at most twice. The only case where a literal appears as often as twice is that a literal with a prime support already on the stack is processed when its variable loses its last implicit support. In this case, the literal must be in zeroLits, and the newly deleted implicit support will be added to the backtrack stack for this literal. But this can only happen once because we delete the literal from zeroLits the first time it happens.

Thus the number of supports is bounded above by $2z$.  □

The bound $2z$ in Lemma 8.1 would improve to $z$ if we maintained zeroLits eagerly instead of lazily, but at the expense of higher overheads elsewhere.

## 9. Experimental Evaluation of HaggisGAC-Stable

We compare HaggisGAC-Stable to HaggisGAC using the same experimental setup as in Section 7. As well as tables of results, we provide a graphical comparison of runtimes of HaggisGAC-Stable and HaggisGAC in Figure 4, and of their memory usage in Figure 5.

Table 4 and Figure 4 shows results for the instances of Section 7.1. We present all four instantiations of HaggisGAC-Stable, along with the fastest instantiation of HaggisGAC, the Watched Element special-purpose propagator, and Constructive Or (which was faster than GAC-Schema in Table 1). For element, we observe about a 10% slowdown, and again a slight slowdown for both List variants. For full-length supports, we see almost identical performance.

Table 5 shows the results for instances of Section 7.2. HaggisGAC-Stable-Lex performs slightly worse than HaggisGAC-Lex, though is in fact never more than 10% worse and very slightly faster on the largest instances. This might be because supports found deep in search are likely to contain more literals than supports found earlier, meaning that when we backtrack the longer supports are retained instead of replaced by the earlier and more efficient short supports. If so, this advantage disappears for the Long variants. Indeed, HaggisGAC-Stable-Long performs much better than HaggisGAC-Long, and the improvement increases with $n$, being about 4.5 times for $n = 24$.

The Rectangle Packing instantiation of ShortGAC described in Section 7 generates an empty support when the constraint becomes entailed, causing all variables to be implicitly supported from that point on. This empty support is not backtrack stable, so cannot be used with HaggisGAC-Stable. We implemented a new backtrack stable variant of findNewSupport, in which the empty support is not returned, but is otherwise the same as before. The List and Long variants are not affected because they do not return the empty support in this case. In Table 6, we use the instances from Section 7.3. Results show significant slowdowns by using backtrack stability for rectangle packing, more than 2 times for $n = 24$. This is probably because of the inability to return the empty support. On the other hand, we see speedups of about 50% for the list variants, and in some cases a factor of 2 speedup for full-length supports.

We see in Figure 5 that the memory usage goes down greatly when stability is used on full-length supports, possibly contributing to speedups in these cases. The greatest reductions are in the case of element, in two cases more than 20 times less memory. On the other hand, there is no significant reduction in memory usage in any non-long variant.

We also tested HaggisGAC-Stable against GAC-Schema as in Section 7.4. This gave very similar performance to HaggisGAC and was therefore better than GAC-Schema: we omit detailed results. There was no significant memory advantage compared to HaggisGAC, with the Stable variant saving less than 25%. We therefore do not seem to gain the advantages we saw earlier from backtrack stability on full-length supports.

We conclude that backtrack stability can speed up HaggisGAC significantly, and greatly reduce memory usage when using full-length supports. However, care must be used, because backtrack stability can be harmful if insisting on backtrack stability increases the size of returned supports.

## 10. Related Work

Our use of counters to count supports is inspired by AC4 (Mohr & Henderson, 1986). There has been some study of compressing the tuples of a constraint into a compact data structure in order to make propagation more efficient. For example, Gent et al. (2007) used tries, and Cheng and Yap (2010) applied MDDs. There has also been extensive study of searching the list of tuples to find the first valid tuple. Approaches include binary search (Lecoutre & Szymanek, 2006), trie search (Gent et al., 2007), and approaches similar to skip lists such as NDLists (Gent et al., 2007) and hologram-tuples (Lhomme, 2004; Lhomme & Régin, 2005). All these techniques are orthogonal to the main focus of this paper because they assist in finding supports, not in maintaining the set of active supports. We have adapted NDLists to contain short supports in Section 5.5.2; it may also be interesting to adapt some of the other approaches.

STR2 maintains a sparse set of all valid satisfying tuples of the constraint (Lecoutre, 2011). Updated variable domains are computed from this set each time the algorithm is invoked. There is no concept of maintaining support, or seeking new support for a literal. It would be interesting to investigate adapting STR2 to handle short supports. This would result in an entirely different algorithm to the ones presented in this paper, possibly with complementary strengths.

| $n$ | WatchElt | HaggisGAC | HaggisGAC-Stable | | | | Con |
|---|---|---|---|---|---|---|---|
| | | Specific | Specific | List | NDList | Long | Or |
| 6 | 27,825 | 11,131 | 10,305 | 4,881 | 2,358 | 30.3 | 53.5 |
| 7 | 22,259 | 9,035 | 8,302 | 4,225 | 1,349 | ‡15.1 | 24.2 |
| 8 | 15,635 | 5,652 | 4,986 | 1,950 | 550 | ‡7.0 | 9.1 |
| 9 | 15,898 | 5,419 | 4,579 | 1,711 | 388 | ‡4.4 | ‡6.2 |
| 10 | 15,088 | ‡4,227 | ‡4,008 | ‡2,409 | ‡309 | ‡2.5 | ‡4.2 |

Table 4: Nodes searched per second for quasigroup existence problems. All columns are named as in Table 1.

| $n$ | GACLex | HaggisGAC | | HaggisGAC-Stable | | GAC- | Con |
|---|---|---|---|---|---|---|---|
| | | Specific | Long | Specific | Long | Schema | Or |
| 3 | 104,955 | 91,265 | 9,288 | 90,473 | 12,008 | 3,622 | 5,735 |
| 4 | 103,950 | 100,100 | 8,628 | 103,470 | 9,056 | 3,030 | 4,997 |
| 5 | 95,420 | 90,009 | 8,503 | 93,382 | 7,248 | 2,734 | 4,104 |
| 6 | 80,841 | 74,184 | 4,666 | 76,777 | 3,844 | 1,638 | 2,109 |
| 7 | 72,307 | 65,359 | 3,271 | 67,273 | 2,615 | 1,190 | 1,188 |
| 8 | 66,445 | 52,659 | 1,609 | 52,113 | 1,591 | 670 | 456 |
| 9 | 64,267 | 47,847 | 914 | 47,881 | 1,114 | 451 | 263 |
| 10 | 57,208 | 39,683 | 634 | 39,176 | 806 | 318 | 184 |
| 12 | 48,146 | 32,425 | 311 | 32,310 | 533 | 170 | 105 |
| 14 | 36,751 | 23,063 | 142 | 23,709 | 345 | 82.3 | ‡99.1 |
| 16 | 30,057 | 18,420 | 90.9 | 18,556 | 248 | 51.5 | ‡62.6 |
| 18 | 22,432 | 13,845 | 53.8 | 14,504 | 177 | 33.3 | ‡48.3 |
| 20 | 16,625 | 10,711 | 38.9 | 10,438 | 135 | 21.0 | ‡36.7 |
| 22 | 12,450 | 8,141 | 26.0 | 8,159 | 106 | 12.5 | ‡27.0 |
| 24 | 9,526 | 6,268 | 18.9 | 6,165 | 85 | ‡7.3 | ‡21.8 |

Table 5: Nodes searched per second for BIBDs. GACLex is the special-purpose propagator for Lex, and all other columns are named as in Table 1.

| $n$-$w$-$h$ | HaggisGAC | HaggisGAC-Stable | | | | GAC- |
|---|---|---|---|---|---|---|
| | Specific | Specific | List | NDList | Long | Schema |
| 18-31-69 | 19,524 | 16,950 | 9,544 | 8,383 | 1,686 | 1,033 |
| 19-47-53 | 39,185 | 22,580 | 4,663 | 8,264 | 1,621 | 1,181 |
| 20-34-85 | 21,000 | 12,865 | 4,950 | 4,840 | 2,607 | 775 |
| 21-38-88 | 12,262 | 9,783 | ‡6,492 | ‡5,827 | 957 | 592 |
| 22-39-98 | 11,966 | 8,798 | ‡4,744 | ‡4,319 | 921 | 518 |
| 23-64-68 | 30,628 | 28,987 | ‡2,377 | ‡4,511 | 1,095 | 590 |
| 24-56-88 | 16,075 | 6,741 | ‡3,894 | ‡3,998 | 1,149 | 474 |
| 25-43-129 | 10,228 | 5,706 | ‡2,405 | ‡2,199 | 1,265 | 348 |
| 26-70-89 | 23,132 | 27,507 | ‡1,689 | ‡4,024 | 890 | 376 |
| 27-47-148 | 4,677 | 3,996 | ‡1,591 | ‡1,735 | 344 | 272 |

Table 6: Nodes searched per second for Rectangle Packing instances. All columns are named as in Table 1.

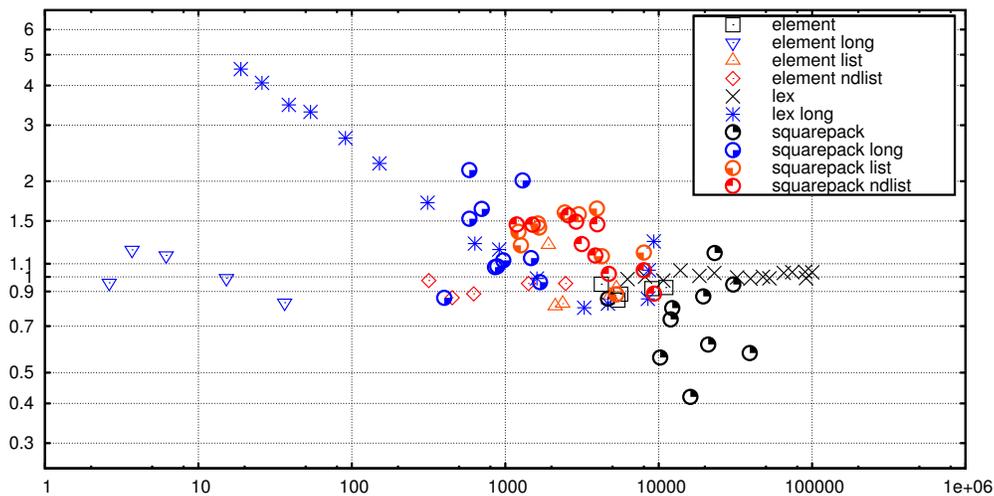

Figure 4: Summary comparison of HAGGISGAC and HAGGISGAC-STABLE. The $x$-axis is median nodes per second for HAGGISGAC. The $y$-axis is speedup (or slowdown) of HAGGISGAC-STABLE.

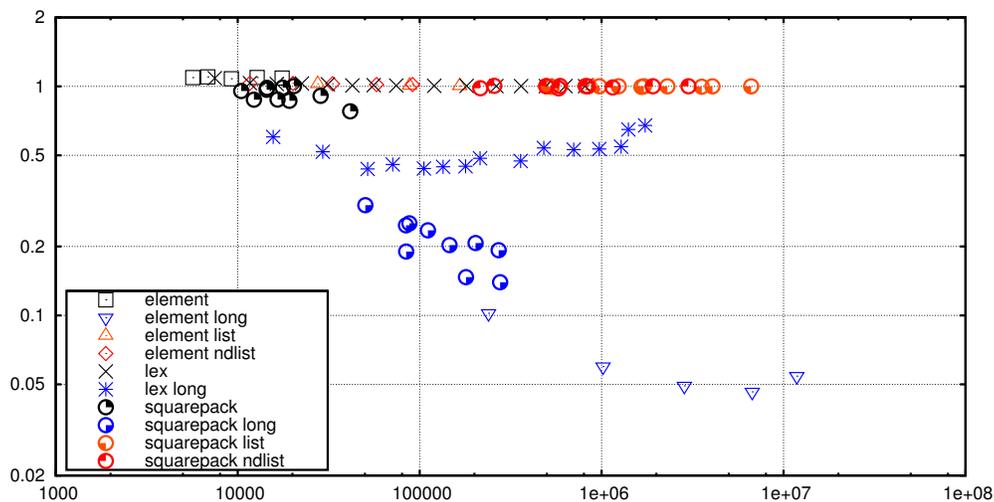

Figure 5: Summary comparison of memory usage (KiB) of HAGGISGAC and HAGGISGAC-STABLE. The $x$-axis is median memory usage for HAGGISGAC. The $y$-axis is reduction (or increase) in usage of HAGGISGAC-STABLE, i.e. the ratio of HAGGISGAC memory usage to that of HAGGISGAC-STABLE. Hence 1 represents equal behaviour, while below 1 means that HAGGISGAC-STABLE used less memory.

The MDD propagator MDDC (Cheng & Yap, 2010) maintains an MDD incrementally during search. The MDD is a compressed representation of the satisfying tuples of the constraint. The time complexity of MDDC is linear in the initial size of the MDD, therefore the degree of compression is vital to the efficiency of the algorithm. In some cases, if a constraint is amenable to strict short supports, it will also compress well into an MDD (given an appropriate variable ordering). For example, the lex constraint compresses well partly because (given the variable order $x_1, y_1, x_2, y_2, \ldots$) the constraint can be satisfied by assigning a prefix of the variables. Lex is amenable to short supports for the same reason. However, some constraints have a small set of short supports but cannot be compressed effectively into an MDD. Suppose we have a disjunction of equality constraints for each pair of $n$ variables of domain size $d$. After $n-1$ variables, the MDD must have $_dC_{n-1}$ states.

Another property of MDD compression might indicate an interesting direction for future work. Lex also compresses well into an MDD because multiple assignments to a prefix of the variables lead to the same subsequent vertex (e.g. $\{x_1 \mapsto 1, y_1 \mapsto 1\}$ and $\{x_1 \mapsto 2, y_1 \mapsto 2\}$). This is something that our short support algorithms are not currently able to exploit.

Katsirelos and Walsh (2007) proposed a different generalisation of support, named $c$-tuples. A $c$-tuple contains a set of values for each variable in the scope of the constraint. Any valid tuple whose values are drawn from the $c$-tuple is a (full-length) support. Katsirelos and Walsh give an outline of a modified version of GAC-Schema which directly stores $c$-tuples. They also present experiments based on a different propagator, GAC3.1r, demonstrating a modest speed improvement for $c$-tuples compared to conventional full-length supports. When a $c$-tuple contains all values of some variable, it will nevertheless be recorded (in $S_C$) as support for each value individually (Katsirelos & Walsh, 2007). The algorithm has no concept of implicit support.

In the context of Constructive Or, Lhomme (2003) observed that a support for one disjunct $A$ will support all values of any variable not contained in $A$. The concept is similar to a short support albeit less general, because the length of the supports is fixed to the length of the disjuncts. He presented a non-incremental Constructive Or algorithm for two disjuncts.

Our algorithms have a similar flavour to GAC-Schema (Bessière & Régin, 1997), so it was natural to compare them to GAC-Schema. However there are other GAC algorithms such as GAC2001/3.1 (Bessière et al., 2005) and it would be interesting to compare these to our algorithms.

## 11. Conclusions

We have introduced and detailed three general purpose propagation algorithms for short supports. They each can either be given a specialised function to find new supports for each constraint, or used with a function that accepts an explicit list of short supports. Where strict short supports are available, all three algorithms perform very well, and provide much better performance than the general purpose methods GAC-Schema and Constructive Or. This shows the value of using strict short supports.

The first algorithm we studied was SHORTGAC, for which we described improvements compared to our earlier report on this algorithm (Nightingale et al., 2011). We identified a significant inefficiency with SHORTGAC when dealing with explicit supports. We introduced a new algorithm, HAGGISGAC which corrects this flaw, has better theoretical complexities, and performs much better than SHORTGAC in our experiments. In three case studies, HAGGISGAC is far faster than the general purpose methods. In the best case it even achieved speeds more than 90% of that of a special purpose propagator. Perhaps remarkably, while able to deal with both strict short and full-length supports, HAGGIS-GAC outperformed SHORTGAC on strict short supports and GAC-Schema on full-length supports, i.e. the cases which those algorithms are respectively specialised for.

Our third algorithm, HAGGISGAC-STABLE, can retain supports on backtracking. It can be less effective than HAGGISGAC if it invalidates the use of certain strict short supports, but it can also be significantly faster on problems with only full-length supports, and reduce memory usage greatly in those cases.

All the proposed algorithms are excellent for propagating disjunctions of constraints. In all experiments with disjunctions we found our algorithms to be faster than Constructive Or and GAC-Schema by at least an order of magnitude, and up to three orders of magnitude.

To summarise, we have shown the value of the explicit use of strict short supports in general purpose propagation algorithms for generalised arc consistency. When strict short supports are available, exploiting them yields orders of magnitude improvements for generic propagation algorithms. In some cases, we even found that a generic algorithm can come close to the performance of a specialised propagator. Previously, short supports do not seem to have been recognised as important in their own right. Our overall contribution is to correct this and focus on short supports as first class objects.

## Acknowledgments

We would like to thank anonymous reviewers and Bilal Syed Hussain for their comments, and EPSRC for funding this work through grants EP/H004092/1 and EP/E030394/1.

## Appendix A. Comparison of SHORTGAC and SHORTGAC-IJCAI

In Section 4, we noted that we have optimised data structures and algorithms for SHORT-GAC, compared with our previous presentation (Nightingale et al., 2011). To demonstrate that these are indeed improvements, we compared the two implementations of SHORTGAC on the three case studies used in this paper. We use the name SHORTGAC-IJCAI for the previous version. We are not quoting results from our previous work (Nightingale et al., 2011), but have rerun all experiments using the environment described in Section 7. We also updated the codebase to Minion 0.12 instead of Minion 0.10 in our earlier paper. For each algorithm and instance, we report nodes searched per second and peak memory use.

Table 7 shows results for the instances of Section 7.1. It is clear from the results that SHORTGAC makes much better use of memory and is also faster than SHORTGAC-IJCAI on these instances. Table 8 shows the results for instances of Section 7.2. As with Element, SHORTGAC makes better use of memory and is faster than SHORTGAC-IJCAI, although improvements are not as great as before. In Table 9, we use the instances from Section 7.3. As in the previous two case studies, SHORTGAC is consistently better in both speed and memory use. We conclude that the algorithms and data structures used in this paper are indeed superior to those we used previously (Nightingale et al., 2011).

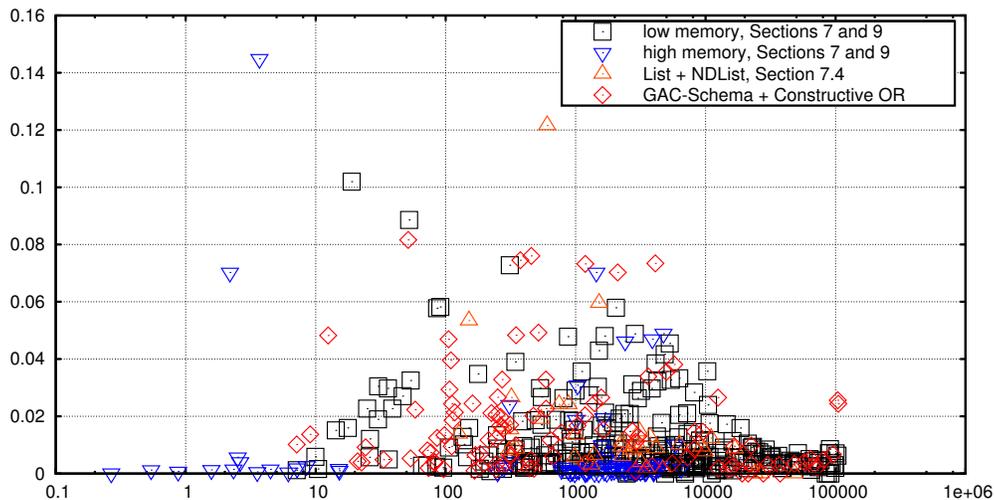

Figure 6: Scatterplot of median nodes per second ($x$-axis) against the median absolute deviation of this divided by the median ($y$-axis). We distinguish between the main experiments of Sections 7 and 9, the cases where medians were of only 5 runs, the list variants used on table constraints in Section 7.4, and data in the paper for GAC-Schema and Constructive Or.

## Appendix B. Median Absolute Deviation of our Experiments

In our experiments we report the median of either 11 or 5 runs. To assess how robust the median was as a measure we looked, for each combination of instances and algorithm, at the *median absolute deviation* (MAD), i.e. the median of the absolute difference of data points from the median. Figure 6 shows the MAD for all algorithm/instance combinations as a fraction of the median for that case. This shows 511 algorithm/instance combinations we tested (including some combinations not reported in detail in this paper). For nodes per second, the maximum MAD we found is always less than 15% of the median, with a worst case of 14.5%. This was HAGGISGAC-Long for $n = 9$ in Table 1. There were only four more cases with MAD above 8% of the median. Figures for memory usage were even more consistent, with only two cases (at 6.3% and 6.1%) showing MAD above 5% of the median and and no others above 2%. Any major conclusions we draw do not regard a 10% change of behaviour between one method and another as significant, and therefore we can say that the median is a robust measure of performance.

| $n$ | ShortGAC node rate | ShortGAC-IJCAI node rate | ShortGAC memory | ShortGAC-IJCAI memory |
|---|---|---|---|---|
| 6 | 6,956 | 4,839 | 5,684 | 27,880 |
| 7 | 4,866 | 3,273 | 6,624 | 72,916 |
| 8 | 2,773 | 1,673 | 8,996 | 188,812 |
| 9 | 2,374 | 1,511 | 12,560 | 461,648 |
| 10 | ‡1,594 | ‡1,294 | ‡17,048 | ‡991,768 |

Table 7: Nodes searched per second and memory use (KiB) for quasigroup existence problems. Comparison of ShortGAC with ShortGAC-IJCAI.

| $n$ | ShortGAC node rate | ShortGAC-IJCAI node rate | ShortGAC memory | ShortGAC-IJCAI memory |
|---|---|---|---|---|
| 3 | 87,463 | 83,964 | 7,476 | 8,392 |
| 4 | 99,602 | 98,135 | 11,680 | 12,992 |
| 5 | 89,127 | 89,286 | 16,408 | 18,512 |
| 6 | 73,260 | 74,184 | 22,568 | 26,260 |
| 7 | 65,062 | 63,091 | 31,348 | 36,356 |
| 8 | 51,335 | 50,480 | 42,420 | 49,012 |
| 9 | 47,059 | 45,085 | 55,660 | 65,684 |
| 10 | 38,344 | 36,179 | 74,348 | 85,700 |
| 12 | 31,626 | 29,455 | 120,024 | 138,496 |
| 14 | 22,712 | 20,868 | 181,252 | 209,492 |
| 16 | 17,813 | 16,087 | 263,792 | 308,400 |
| 18 | 13,843 | 12,356 | 360,500 | 422,536 |
| 20 | 10,734 | 9,614 | 493,368 | 570,188 |
| 22 | 7,976 | 7,208 | 632,064 | 735,548 |
| 24 | 6,255 | 5,398 | 811,104 | 939,796 |

Table 8: Nodes searched per second and memory use for BIBD problems. Comparison of ShortGAC with ShortGAC-IJCAI.

| $n$-$w$-$h$ | ShortGAC node rate | ShortGAC-IJCAI node rate | ShortGAC memory | ShortGAC-IJCAI memory |
|---|---|---|---|---|
| 18-31-69 | 14,923 | 10,892 | 11,876 | 24,568 |
| 19-47-53 | 38,329 | 29,647 | 10,172 | 19,680 |
| 20-34-85 | 13,949 | 10,288 | 13,988 | 33,020 |
| 21-38-88 | 8,568 | 6,109 | 16,100 | 38,828 |
| 22-39-98 | 8,059 | 5,821 | 18,868 | 46,344 |
| 23-64-68 | 31,486 | 24,528 | 13,988 | 31,700 |
| 24-56-88 | 12,317 | 8,386 | 17,548 | 43,708 |
| 25-43-129 | 5,310 | 3,828 | 27,580 | 74,064 |
| 26-70-89 | 25,860 | 21,146 | 19,796 | 49,512 |
| 27-47-148 | 2,943 | 2,086 | 39,848 | 106,144 |

Table 9: Nodes searched per second and memory use for rectangle packing. Comparison of ShortGAC with ShortGAC-IJCAI.

## Appendix C. Comparison of GAC-Schema and HaggisGAC

We showed in Section 7.4 that HaggisGAC outperforms GAC-Schema when dealing with full-length supports. This is despite the fact that HaggisGAC has small overheads for dealing with strict short supports even when none exist. We now discuss briefly why this may be so.

GAC-Schema has the concept of *current* supports – each literal has one current support, which is one of the active supports that contain the literal. There is an additional data structure $S(\tau)$. For each active support $\tau$, $S(\tau)$ is a list of all literals that have $\tau$ as their current support. Hence when $\tau$ is invalidated, GAC-Schema finds a new current support for each literal in $S(\tau)$ (or deletes the literal). In HaggisGAC we dispensed with this entirely. The sign that a literal needs a new support is not that it lost its *current* support, but that its support list (supportListPerLit) is empty. There is a small potential saving from not maintaining $S(\tau)$.

A second, possibly more important, difference is that GAC-Schema is more eager than HaggisGAC. When a literal $x \mapsto v$ loses its current support, GAC-Schema will check if other active supports containing $x \mapsto v$ are valid, an $O(n)$ operation for each one. If they are all invalid, GAC-Schema then calls findNewSupport. If this returns Null then $x \mapsto v$ is deleted. HaggisGAC does none of this, avoiding completely the cost of checking validity. This is safe because if every support is invalid, the literal deletion from each support will cause a call to deleteSupport and the last will result in the empty list, causing a call to findNewSupport. Both approaches are correct, but GAC-Schema's is wasteful because it performs unnecessary validity checks. However, one cannot guarantee time saving, because GAC-Schema can perform deletions sooner, possibly affecting the way the propagator interacts with the other propagators.